\documentclass[twoside]{article}

\usepackage[accepted]{aistats2025}
\usepackage[round]{natbib}

\usepackage{amsmath,amsfonts,bm}

\def\eqref#1{equation~\ref{#1}}

\def\1{\bm{1}}

\def\rmC{{\mathbf{C}}}

\DeclareMathAlphabet{\mathsfit}{\encodingdefault}{\sfdefault}{m}{sl}
\SetMathAlphabet{\mathsfit}{bold}{\encodingdefault}{\sfdefault}{bx}{n}

\def\gP{{\mathcal{P}}}

\def\gX{{\mathcal{X}}}
\def\gY{{\mathcal{Y}}}

\usepackage{hyperref}
\usepackage{url} 

\usepackage{booktabs}
\usepackage{amsmath}
\usepackage{amssymb}
\usepackage{mathtools}
\usepackage{amsthm}

\usepackage{wrapfig}
\usepackage{import}
\usepackage{graphicx}
\usepackage{todonotes} 
\usepackage[utf8]{inputenc}
\usepackage[english]{babel} 
\usepackage{dsfont}
\usepackage{lipsum}  
\usepackage[mathscr]{euscript}
\usepackage[normalem]{ulem}
\usepackage{enumitem}

\usepackage{caption}
\usepackage{subcaption}

\theoremstyle{plain}
\newtheorem{theorem}{Theorem}[section]
\newtheorem{proposition}[theorem]{Proposition}
\newtheorem{lemma}[theorem]{Lemma}
\newtheorem{corollary}[theorem]{Corollary}
\theoremstyle{definition}

\theoremstyle{remark}

\usepackage{tcolorbox}

\usepackage[rightcaption]{sidecap}

\usepackage{multirow}
\usepackage[capitalize,noabbrev]{cleveref}

\begin{document}

%

%

\twocolumn[

\aistatstitle{Task-Driven Discrete Representation Learning}

\aistatsauthor{ Tung-Long Vuong}

\aistatsaddress{Monash University} ]

\begin{abstract}

In recent years, deep discrete representation learning (DRL) has achieved significant success across various domains. Most DRL frameworks (e.g., the widely used VQ-VAE and its variants) have primarily focused on generative settings, where the quality of a representation is implicitly gauged by the fidelity of its generation. In fact, the goodness of a discrete representation remain ambiguously defined across the literature. In this work, we adopt a practical approach that examines DRL from a task-driven perspective. We propose a unified framework that explores the usefulness of discrete features in relation to downstream tasks, with generation naturally viewed as one possible application. In this context, the properties of discrete representations as well as the way they benefit certain tasks are also relatively understudied. We therefore provide an additional theoretical analysis of the trade-off between representational capacity and sample complexity, shedding light on how discrete representation utilization impacts task performance. Finally, we demonstrate the flexibility and effectiveness of our framework across diverse applications. 


\end{abstract}

\section{Introduction}

Deep neural representation learning has been a key factor in the success of deep learning across various domains, including images \citep{pathak2016context, goodfellow2014generative, kingma2016improved}, videos, and audio \citep{reed2017parallel, oord2016wavenet, kalchbrenner2017video}. The fundamental idea is to use deep neural networks to learn compact, expressive representations from high-dimensional data, typically represented as continuous-valued vectors \citep{kingma2013auto, chen2020simple, chen2021exploring}. However, for certain tasks such as complex reasoning, planning, or predictive learning \citep{hafner2020mastering, allen2021learning, ozair2021vector, van2017neural}, it can be more advantageous to interpret non-identical but similar vectors as internal representations of shared discrete symbols from a fixed codebook, leading to the concept of discrete representation learning \citep{ozair2021vector, hu2017learning}.

In response, discrete representations have been actively explored and applied in various contexts. The pioneering, also most popular, unsupervised discrete representation learning framework is \textit{Vector Quantization Variational Auto-Encoder} (VQ-VAE) \citep{van2017neural}, followed by its variants \citep{roy2018theory,takida22a,yu2021vector,razavi2019generating, williams2020hierarchical,vuong2023vector}. \citep{van2017neural} successfully extends VAE model for learning discrete latent representations. Concretely, the vector quantized models learn a compact discrete representation using a deterministic encoder-decoder architecture in the first stage, and subsequently apply this highly compressed representation for various downstream tasks, namely image generation \citep{esser2021taming}, cross-modal translation \citep{kim2022verse}, image recognition \citep{yu2021vector} or text-to-image generation \citep{ramesh2021zero,rombach2022high}. Another line of research which can be categorized as discrete representation learning is \textit{Prototypical Representation Learning} \citep{yarats2021reinforcement, ming2022exploit, lin2023frequency, ye2024ptarl}, where discrete representations are used to model clusters, classes, or patterns in data in a supervised fashion. The role of prototypes varies across tasks. \citet{ming2022exploit} use prototypes for out-of-distribution (OOD) detection and data points are examined through their relation to these representations;  prototypes are also considered as summaries, for instance, of an agent’s exploratory experiences \citep {yarats2021reinforcement} or of continuous representations for classification tasks \citep{lin2023frequency}. The learning objectives also vary based on differences in usage. For example, \citet{islam2022representation} encourage the alignment of augmented data representations with the original data while maximizing the information-theoretic dependency between the data and their predicted discrete representations. Alternatively,  \citet{liu2022adaptive} refines the prototype learning process by employing  
discrete information bottlenecks for learning structured representation spaces. From a prototypical perspective, any classification problem in fact can be viewed as a form of discrete representation learning, where continuous representations are classified based on their similarity to the class-specific vectors. This effectively transforms continuous features into discrete class labels by matching them to the most similar prototype.

Although deep discrete representation learning has been widely applied across various domains, the learning processes for discrete representations vary significantly between tasks, making it difficult to analyze their properties and advantages comprehensively. Much of the existing research on discrete representations focuses primarily on VQ-VAE and its variants. For instance, works such as \citep{roy2018theory, takida22a, yu2021vector, Zheng2022MOvq, dhariwal2020jukebox, williams2020hierarchical} explore issues like codebook collapse and propose solutions to mitigate them. This highlights the need for a generalized framework that encapsulates discrete representation learning, along with theoretical analyses to better understand its properties.

\textbf{Contribution.} In this work, we introduce a generalized framework for learning task-driven deep discrete representations. We present a theoretical analysis on learned discrete representations, highlighting the trade-off between optimal task accuracy and sample complexity. Specifically, reducing the size of the codebook decreases the optimal accuracy but requires fewer samples to achieve generalization. On the other hand, increasing the codebook size improves accuracy but necessitates a larger sample size. We translate the theoretical results into practical insights about how to effectively optimize discrete latents in specific downstream tasks. We conduct various experiments to demonstrate the flexibility and generalizability of our framework where it achieves competitive performance across diverse applications.

    


\paragraph{Organization.} The paper is structured as follows. First, we provide a general overview of representation learning and introduce the background on Wasserstein distance, which serves as the foundation for our proposed framework. In Section 3.1, we present the discrete representation framework and derive a trainable form compatible with gradient backpropagation frameworks such as PyTorch. Section 3.2 contains a detailed analysis of our framework. Next, we conduct experiments in Section 4 to empirically validate our theoretical findings, followed by concluding remarks in Section 5. Additional results and detailed proofs are provided in the supplementary material.



    
    



\section{Preliminaries}

\paragraph{Notations.} We first introduce the notations and basic concepts in the paper. We use calligraphic letters (i.e., $\mathcal{X}$) for spaces, upper case letters (i.e. $X$) for random variables, lower case letters (i.e. $x$) for their values and $\mathbb{P}$ for (observed) probability distributions. We consider the task-driven setting where the observation distribution is $\mathbb{P}(x)$ (or $\mathbb{P}_x$) and labelling function $f_y(\cdot)$ is a deterministic mapping from $x$ to the outcomes/labels the task of interest (e.g., a categorical label for a classification problem, a real value/vector for a regression problem).



\paragraph{General Representation Learning.} 

The objective of representation learning is to learn the \emph{encoder} $f_{e}: \mathcal{X}\rightarrow \mathcal{Z}$ first map the data examples $x\in \mathcal{X}$ to the latent $z\in \mathcal{Z}$. The latent representation $z=f_e(z)$ is expected to encapsulate task-related information from $x$ and is then used to predict the task $f_y(x)$ using a decoder $f_d$. $f_e, f_d$ usually is learned by the following optimization problem (OP): 
\begin{align}
\min_{f_{d},f_{e}}\mathbb{E}_{x\sim\mathbb{P}_{x}}\left[\ell_{y}\left(f_{d}\left(f_{e}\left(x\right)\right),f_y(x)\right)\right]
\label{eq:simple_ob}
\end{align}


where $\ell_y(u,v)$ is the loss between $u$ and $v$ which is depended on our task of interest. For instance, $\ell_y$ is cross-entropy loss for classification task, L1 or L2 loss for regression task.

\paragraph{Optimal transport.}  Let $\mu = \sum^{n}_{i=1} a_i \delta_{x_i}$ and  $\nu = \sum^{m}_{j=1} b_j \delta_{y_j}$
are discrete probability distributions where $a_j$ and $b_j$ represent discrete probability masses such that $\sum_{i=1}^n a_i=1$ and $\sum_{j=1}^m b_j=1$. The Kantorovich formulation of the OT distance \citep{kantorovich2006problem} between two discrete distributions $\alpha$ and $\beta$ is 
\begin{equation}\label{eq:kanto_disc}
    W_c \left(\mu, \nu \right) := \inf_{\mathbb{P} \in \Pi(\mu,\nu)} \langle \rmC, \mathbb{P} \rangle,
\end{equation}
where $\langle \cdot, \cdot \rangle$ denotes the Frobenius dot-product; $\rmC \in \mathbb{R}^{n \times m}_{+}$ is the cost matrix of the transport; $\mathbb{P} \in \mathbb{R}^{n \times m}_{+}$ is the transport matrix/plan; $\Pi(\mu,\nu) := \left\{ \mathbb{P} \in \mathbb{R}^{n \times n}_{+} : \mathbb{P} \mathbf{1}_{n} = \mu, \mathbb{P} \mathbf{1}_{n} = \nu \right\} $ is the transport polytope of $\mu$ and $\nu$; $\mathbf{1}_{n}$ is the $n-$dimensional column of vector of ones. For arbitrary measures, Eq. (\ref{eq:kanto_disc}) can be generalized as 
\begin{equation}\label{eq:kanto_cont}
    W_c \left(\mu; \nu\right) := \underset{\pi \sim \mathcal{P} (X \sim \mu, Y \sim \nu)}{\mathrm{inf}} \mathbb{E}_{(x, y) \sim \pi} \bigl[ c(x,y)\bigr],
\end{equation}
where the infimum is taken over the set of all joint distributions $\gP (X \sim \alpha, Y \sim \beta)$ with marginals $\alpha$ and $\beta$ respectively. $c: \gX \times \gY \mapsto \mathbb{R}$ is any measurable cost function. If $c(x,y) = D^p(x,y)$ is a distance for $p \le 1$, $W_p$, the $p$-root of $W_c$, is called the $p$-Wasserstein distance.  Finally, for a measurable map $T : \gX \mapsto \gY$, $T\#\mu$ denotes the push-forward measure of $\mu$ that, for any measurable set $B \subset \gY$, satisfies $T\#\mu(B) = \mu(T^{-1}(B))$. For discrete measures, the push-forward operation is $T\#\mu = \sum^{n}_{j=1} a_j \delta_{T(x_j)}$.

\section{Discrete Representation Learning}
\begin{figure}[h!]
\begin{centering}
\subfloat{\centering{}\includegraphics[width=0.8\linewidth]{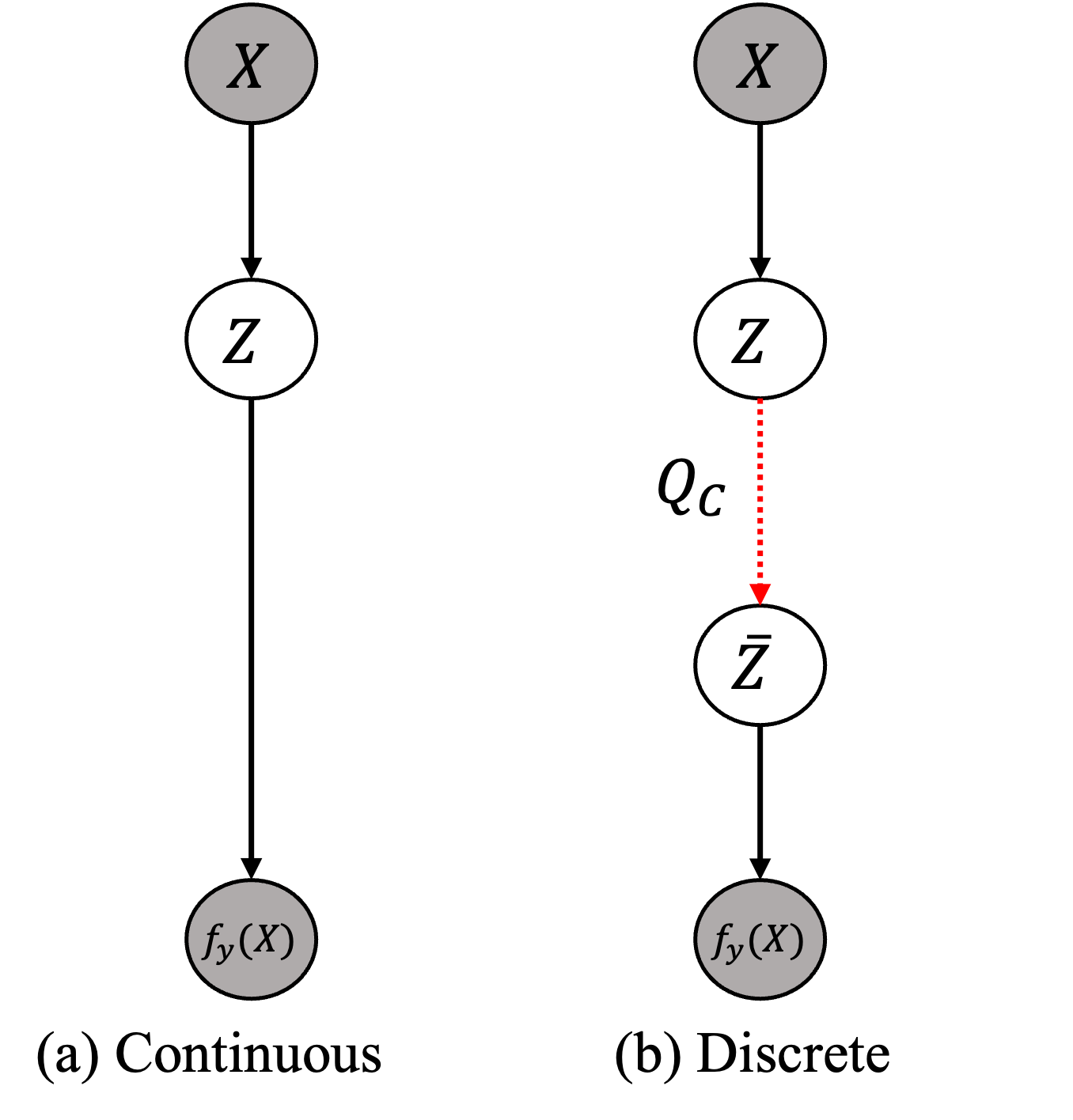}}
\par\end{centering}
\caption{Graphical model: (a) continuous representation vs (b) discrete representation where discrete representation $\bar{z}$ is constrainted by set of $M$ discrete representation i.e.,  $\bar{Z}\in C=\{c_1,...,c_M\}.$}
\label{fig:DG}
\end{figure}

\textbf{Codebook.} Unlike continuous representation learning, in the discrete representation learning (DRL) framework, the image of $f_e$ is constrained to a finite set of discrete representations, commonly referred to as the \emph{codebook}.
The \emph{codebook} consists of a set of codewords $C=\{c_{m}\}_{m=1}^{M}$, where each codeword belongs to the latent space $\mathcal{Z}$ ($C\in \mathcal{Z}^{M}$). 

\textbf{Quantization.} Denote the discrete representation function $\bar{f}_e=Q_C \circ f_e$ in which the \emph{encoder} $f_{e}: \mathcal{X}\rightarrow \mathcal{Z}$ first map the data examples $x\in \mathcal{X}$ to the latent $z\in \mathcal{Z}$. Following, a quantization $Q_C$ projecting $z$ onto $C: \bar{z} =Q_C(z) = \underset{c\in C}{\text{argmin}}d_{z}\left(z,c\right)$ is a quantization operator which $d_z$ is a distance on $\mathcal{Z}$.

\textbf{Objective function.} Since the image of $\bar{f}_e$ is constrainted by $C$, from probabilistic viewpoint, we can endow discrete distribution $\mathbb{P}_{c,\pi}= \sum_{m=1}^{M}\pi_{m}\delta_{c_{m}}$ over the codebook $C$ where the category probabilities $\pi \in \Delta_{M} = \{\alpha \ge 0 : \Vert \alpha \Vert_1 = 1 \}$ lie in the $M$-simplex. Then $\bar{f}_{e}: \mathcal{X} \mapsto \mathcal{Z}$ be a deterministic encoder such that $\bar{f}_{e}\#\mathbb{P}_x = \mathbb{P}_{c,\pi}$. The OP of discrete representation learning framework w.r.t training data $\mathbb{P}_x$ can be formulated as the minimizer of following OP:
\begin{align}
\min_{\mathbb{P}_{c,\pi},f_{d}}\min_{\bar{f}_{e}:\bar{f}_{e}\#\mathbb{P}_{x}=\mathbb{P}_{c,\pi}}   \underset{x\sim\mathbb{P}_{x}}{\mathbb{E}}\left[\ell_{y}\left(f_{d}\left(\bar{f}_{e}\left(x\right)\right),f_y(x)\right)\right] 
\label{eq:single_level}
\end{align}

This OP explicitly impose the constraint on $\bar{f}_e$ i.e., $\bar{f}_{e}$ is a deterministic discrete encoder mapping a data example $x$ directly to a codeword which prevent us from end-to-end training using gradient backprobagation. To make it trainable,
we replace $\bar{f}_{e}$ by a continuous encoder $f_{e}:\mathcal{X}\rightarrow\mathcal{Z}$ in the following theorem.

\begin{theorem}
\label{thm:trainable_single}
If we seek $f_{d}$ and $f_{e}$ in a family
with infinite capacity (e.g., the family of all measurable functions),
the the two OPs of interest in (\ref{eq:single_level}) is equivalent to the following OPs:
\begin{align}
\min_{\mathbb{P}_{c,\pi}} \min_{f_{d},f_{e}}&
\mathbb{E}_{x\sim\mathbb{P}_{x}}\left[\ell_{y}\left(f_{d}\left(Q_C\left(f_{e}\left(x\right)\right)\right),f_y(x)\right)\right]\nonumber\\
&+\lambda\mathcal{W}_{d_{z}}\left(f_{e}\#\mathbb{P}_{x},\mathbb{P}_{c,\pi}\right),
\label{eq:reconstruct_form_continuous}
\end{align}

where $\mathcal{W}_{d_{z}}\left(\cdot,\cdot\right)$ is wasserstein distance with $d_z$ is the metric on representation space, the parameter $\lambda>0$ and $Q_C(z)  = \text{argmin}{}_{c\in C}d_{z}\left(z,c\right)$.
\end{theorem}

This theorem transforms the constrained optimization problem in Eq. (\ref{eq:single_level}) into the unconstrained optimization problem in Eq. (\ref{eq:reconstruct_form_continuous}), which involves task performance $f_d(\cdot)$ from the codeword $Q_C\left(f_{e}\left(x\right)\right)$ and regularization using the Wasserstein distance on the latent space.
We note that although this theorem requires the function classes of $f_e$ and $f_d$ to have infinite capacity, practical considerations necessitate that we restrict ourselves to classes of finite capacity for learning to be feasible. Therefore, we interpret (5) as a practical realization that converges to (4) under ideal conditions.

\subsection{Theoretical Analysis}

In the previous section, we introduced a general framework for discrete representation learning, where the output of the representation function is constrained by the codebook $C=\{c_m\}_{m=1}^M$. The size of this codebook plays a crucial role in DRL framework, as it controls the expressiveness of the representations. To provide a more comprehensive understanding of OP in Eq. (\ref{eq:single_level}), we examine how the codebook size influences the performance upper-bound in Eq.(\ref{eq:single_level}), assuming an infinite data scenario.
Furthermore, we investigate the sample complexity, specifically determining the number of samples required to achieve a low approximation error for the performance upper-bound. This analysis will be discussed in detail in the subsequent section.

\subsubsection{Accuracy Trade-off}

We would like to examine how the codebook size $M$ affects the expected loss in Eq. (\ref{eq:single_level}), as outlined in the following Lemma:

\begin{lemma}
\label{cor:distortion_data} Let $C^{*}=\left\{ c_{m}^{*}\right\} _{k=1}^M,\pi^{*}$,
and $f_{d}^{*}$ be the optimal solution of the OP in Eq. (\ref{eq:single_level}), then $C^{*}=\left\{ c_{m}^{*}\right\} _{k},\pi^{*}$,
and $f_{d}^{*}$ are also the optimal solution of the following OP:
\begin{equation}
\epsilon^{*}_{M} = \min_{f_{d}}\min_{\pi}\min_{\sigma\in\Sigma_{\pi}}\mathbb{E}_{x\sim \mathbb{P}_x}\left [\ell_{y}\left(f_y(x),f_{d}\left(c_{\sigma(x)}\right)\right) \right ]
\nonumber
\end{equation}
and $$\epsilon^{*}_M \geq \epsilon^{*}_{M+1} \text{ } \forall M$$ 
and $$\epsilon^{*}_M \rightarrow 0 \text{ as } M\rightarrow\infty$$ 
where $\Sigma_{\pi}$ is the set of assignment functions $\sigma:\mathcal{X} \rightarrow\left\{ 1,...,M\right\} $
such that $ \mathbb{P}_x\left(\sigma^{-1}\left(m\right)\right),m=1,...,M$
are proportional to $\pi_{k},m=1,...,M$  (e.g., $\sigma$ is the nearest assignment: $\sigma^{-1}\left(m\right) = \{x\mid \bar{f_{e}}\left(x\right) = c_m, m=\text{argmin} _{m}d_{z}\left(f_{e}\left(x\right),c_{m}\right)\}$ is set of latent representations which are quantized to $m^{th}$ codeword).
\end{lemma}

Lemma \ref{cor:distortion_data} states that for the optimal solution
$C^{*}=\left\{ c_{k}^{*}\right\}_{m=1}^M ,\pi^{*}$, $f_e^*$, and $f_{d}^{*}$ of the
OP in (\ref{eq:single_level}), each $f_y(x)$ is assigned to the centroid $f_d^*(c^{*}_{\sigma^{*}}(x))=f_d^*(Q_C(f_e^{*}(x)))$ which forms optimal clustering centroids of the optimal clustering solution minimizing the loss of interest. This lemma also demonstrates that constraining the image of the representation function by the codebook $C=\{c_m\}_{m=1}^M$ leads to a decrease in performance. The smaller the value of $M$ the lower the optimal accuracy that can be achieved.

\textit{Remark.1} \textbf{(Multi-level Discrete Representation)} Real-world data often exhibits a natural hierarchical structure, where data can be grouped at varying levels of granularity. Lemma~\ref{cor:distortion_data} ensures that we obtain the best description of the data given a predefined codebook size, and this description becomes more detailed as the codebook size increases. This enables the construction of multi-level discrete representations, allowing us to explore the data at different levels of detail. For instance, with a small codebook size, you can analyze global patterns, while a larger codebook size allows you to zoom in to examine finer groupings and more specific relationships within the data.

\subsubsection{Sample Complexity} 
Given a specific codebook size $M$ and data distribution $\mathbb{P}_x$, solving OP in Eq. (\ref{eq:single_level})  provides the optimal accuracy corresponding to $M$ and $\mathbb{P}_x$. However, in practice, models are often trained on a finite set of samples drawn i.i.d. from an underlying distribution. We wish that models trained on finite samples can perform well even on previously unseen samples from the underlying distribution. Therefore, we focus on analyzing the gap between the optimal solution obtained from the training data and its performance on both the training data and the underlying distribution. 

Let the training set be $S=\left\{ x_{1},...,x_{N}\right\}$ consisting of $N$ sample $x_i\sim \mathbb{P}_x$ and define the empirical measure $\mathbb{P}^N_{x}$ by placing mass $N^{-1}$ at each of members of $S$ i.e., $\mathbb{P}^N_{x}= \sum_{i=1}^{N}\frac{1}{N}\delta_{x_{i}}$.
Given $C^{*}=\left\{ c_{m}^{*}\right\}_{m=1}^M ,\pi^{*}$, $f_e^*$, and $f_{d}^{*}$ be the optimal solution of the OP in Eq. (\ref{eq:single_level_apd}) w.r.t to empirical training data $\mathbb{P}^N_{x}$, the empirical loss w.r.t to distribution $\mathbb{P}^N_x$ is defined as:
\begin{equation}
\mathcal{L}\left(\mathbb{P}^N_x\right) =\mathbb{E}_{x\sim \mathbb{P}^N_x}\left [ \ell\left(f_y(x),f_d^*(Q_C(f_e^{*}(x)))\right)\right ],\label{eq:true_loss}
\end{equation}
Notably, Lemma \ref{cor:distortion_data} establishes that solving the optimization problem (OP) in Eq. (\ref{eq:single_level}) is equivalent to solving a clustering problem in the output space. However, discrete representations exhibit distinct characteristics. Specifically, defining $\bar{f}_y(x)=f_d^*(\underset{c''\in C}{\text{argmin}}\ell_y(x, f^*_d(c'')))$, we obtain:
\begin{align}
&\mathcal{L}\left(\mathbb{P}^N_x\right) =\mathbb{E}_{x\sim \mathbb{P}^N_x}\left [ \ell\left(f_y(x),f_d^*(\bar{f}_e(x))\right)\right ]\nonumber\\
=& \underset{\text{Clustering Loss: }\mathcal{L}_C\left(\mathbb{P}^N_x\right)}{\underbrace{\mathbb{E}_{x\sim \mathbb{P}^N_x}\left [\ell\left(\bar{f}_y(x),f_y(x)\right)\right ]}}\nonumber\\
&+ \underset{\text{Assignment Discrepancy Loss: }\mathcal{L}_A\left(\mathbb{P}^N_x\right)}{\underbrace{\mathbb{E}_{x\sim \mathbb{P}^N_x}\left [\ell\left(f_{d}^*\left( (\bar{f}_{e}^*\left(x\right) \right),f_y(x)\right) -\ell\left(\bar{f}_y(x),f_y(x)\right)\right ]}}\nonumber\\
=&\mathcal{L}_C\left(\mathbb{P}^N_x\right)+\mathcal{L}_A\left(\mathbb{P}^N_x\right) \nonumber
\end{align}
The overall error is decomposed into two components:
\textit{Clustering Loss}: The fist term $\ell\left(\bar{f}_y(x),f_y(x)\right)$ represents the distance between the label $f_y(x)$ and its nearest centroid in the output space, which is determined by mapping the codebook to the output space.

\begin{figure}[h!]
\begin{centering}
\subfloat{\centering{}\includegraphics[width=1.0\linewidth]{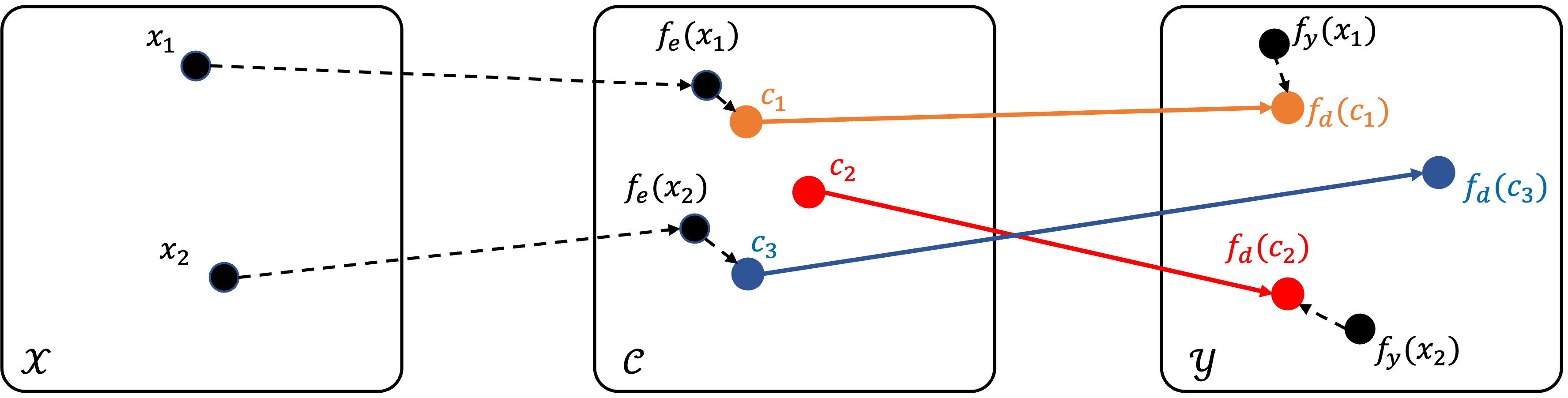}}
\par\end{centering}
\caption{Mismatching between the nearest assignments in the latent and output spaces.}
\label{fig:assignment}
\end{figure}

\textit{Assignment Discrepancy Loss}: It can be seen that $\bar f_e^*(x)$ is the nearest assignment on latent space while $\bar{f}_y(x)=f_d^*(\underset{c''\in C}{\text{argmin}}\ell_y(x, f^*_d(c'')))$ is the nearest assignment on output space. The second term i.e., the discrepancy between $\ell\left(f_{d}^*\left( (\bar{f}_{e}^*\left(x\right) \right),f_y(x)\right) -\ell\left(\bar{f}_y(x),f_y(x)\right)$ arises from the mismatch between assignments in the latent and output spaces. Specifically, in Figure \ref{fig:assignment}, for sample 
, the decoded centroid $f_d(c_3)$ in the latent space may not be the closest centroid in the output space.


Following that, the general loss is defined similarly:
\begin{align}
\mathcal{L}\left(\mathbb{P}_x\right) =\mathcal{L}_C\left(\mathbb{P}_x\right)+\mathcal{L}_A\left(\mathbb{P}_x\right) \nonumber
\end{align}
The discrepancy between the empirical loss and the general loss can be bounded as follows:
\begin{align}
&\left |\mathcal{L}\left(\mathbb{P}_x^N\right) - \mathcal{L}\left(\mathbb{P}_x\right) \right |\nonumber\\
&= \left | \mathcal{L}_C\left(\mathbb{P}^N_x\right)+\mathcal{L}_A\left(\mathbb{P}^N_x\right) - \mathcal{L}_C\left(\mathbb{P}_x\right)-\mathcal{L}_A\left(\mathbb{P}_x\right) \right |\nonumber\\
&\leq \left | \mathcal{L}_C\left(\mathbb{P}^N_x\right)-\mathcal{L}_C\left(\mathbb{P}_x\right)\right | +\left |\mathcal{L}_A\left(\mathbb{P}^N_x\right)-\mathcal{L}_A\left(\mathbb{P}_x\right) \right |\nonumber
\end{align}
The first term $\left | \mathcal{L}_C\left(\mathbb{P}^N_x\right)-\mathcal{L}_C\left(\mathbb{P}_x\right)\right |$ is the generalization bounds for k-mean problem, which is studied in ~\citep{telgarsky2013moment, bachem2017uniform}. In this study, our focus is on the second term, $ \left |\mathcal{L}_A\left(\mathbb{P}^N_x\right)-\mathcal{L}_A\left(\mathbb{P}_x\right) \right |$ , which examines the impact of discrete representations on the optimization problem.

\begin{corollary}
\label{cor:sample_complexity}
    Given a dataset of size $N$, a loss function $\ell_y$ that is a proper metric and upper-bounded by a positive constant $L$, and an encoder class $\mathcal{F}_e$ (i.e., $\bar{f}_e\in\mathcal{F}_e$), then it holds with probability at least $1-\delta$ that
    \begin{equation}
        |\mathcal{L}_A\left(\mathbb{P}_x^N\right) - \mathcal{L}_A\left(\mathbb{P}_x\right)| \leq \epsilon
    \end{equation}
where $\epsilon = L{M}{\frac{\sqrt{\log 1/\delta + \log |\mathcal{F}_e|} + 1}{\sqrt{N}}} $.
\end{corollary}

This bound shows that, given $C^{*}=\left\{ c_{m}^{*}\right\}_{m=1}^M, \pi^{*}$, $\sigma*$, and $f_{d}^{*}$ as the optimal solution to the optimization problem in Eq. (\ref{eq:single_level}) with respect to the empirical training data $\mathbb{P}^N_{x}$, and apply this solution to ground-truth underlying distribution $\mathbb{P}_x$, the number of samples 
$N$ required to to achieve a generalization error of at most $\epsilon$ with high probability $1-\delta$ is given by: $N(\epsilon,\delta)=\Omega\left( \frac{L^2 M^2}{\epsilon^2} \max\left( \log|\mathcal E|/\delta, 1 \right) \right)$. This analysis indicates that the convergence rate of $\left |\mathcal{L}_A\left(\mathbb{P}^N_x\right)-\mathcal{L}_A\left(\mathbb{P}_x\right) \right |$ is $\mathcal{O}(N^{-1/2})$ , which aligns with the convergence rate of the clustering generalization error $\left |\mathcal{L}_C\left(\mathbb{P}^N_x\right)-\mathcal{L}_C\left(\mathbb{P}_x\right) \right |$ as established in \citep{bachem2017uniform}. Consequently, the overall convergence rate of $\left |\mathcal{L}\left(\mathbb{P}^N_x\right)-\mathcal{L}\left(\mathbb{P}_x\right) \right |$ also follows $\mathcal{O}(N^{-1/2})$. 

Moreover, the sample complexity $N(\epsilon, \delta)$ is influenced by the codebook size $M$, scaling as $\Omega(M^2)$, whereas in \citep{bachem2017uniform}, it scales as $\Omega(M\log M)$ in \citep{bachem2017uniform}. Therefore, the sample complexity for the original loss $\left |\mathcal{L}\left(\mathbb{P}^N_x\right)-\mathcal{L}\left(\mathbb{P}_x\right) \right |$ is also $\Omega(M^2)$. This implies that when $M$ is small, fewer samples are required for the model to generalize effectively, whereas larger values of $M$ demand a significantly higher number of samples.


\textit{Remark.2} \textbf{(Sample Complexity Trade-off)} From the two analyses above, we can observe a trade-off between optimal accuracy and sample complexity. Specifically, reducing the size of the codebook decreases the optimal accuracy, but requires fewer samples to guarantee the generalization. Conversely, increasing the codebook size improves accuracy but demands more samples. The trade-off give us an hint choose appropriate codebook size depend on our task of interest. Specifically, for some that that not require very detail representation, we can use small codebook size to gain advantage from small sample complexity.
\section{Experiments}

In this section, we conduct experiments to validate two key properties of our framework. Specifically, we focus on: (i) \textbf{sample complexity trade-off and performance Trade-off} (Section~\ref{sec:RL}): we apply state abstraction problem in reinforcement learning (RL) setting as a natural use case of discrete representation in RL domains to showcase the  trade-off properties. By aggregating states into the same abstract states, similar action distributions are produced, enabling abstract states to be parameterized as discrete representations. We show that discrete-state representations can improve the sample efficiency of RL agents, requiring fewer samples to achieve comparable performance; (ii) \textbf{The multi-level discrete representation} (Section~\ref{sec:DG}): we explore the representation-alignment problem in the domain generalization (DG) setting, where the goal is to align representation distributions across different domains to learn representations that are invariant to domain shifts. Our experiments demonstrate that employing multi-level discrete representations enhances the representation alignment process, leading to improved performance.

\subsection{State Abstraction RL}
\label{sec:RL}
\textbf{Setting.} 
We consider the RL environment characterized by a Block Markov Decision Process (MDP) defined by the tuple  $\langle \mathcal{X}, \mathcal{A}, \mathcal{R}, \mathcal{P}, \gamma \rangle$, representing the set of states, actions, reward function, transition functions, and discount factor, respectively. Block MDP assumes that a true, hidden state space $\mathcal{S}$ exists with a much smaller and more compact representation than the observed states in $\mathcal{X}$.
In state abstraction RL, we aim to learn an abstraction function $\phi$ that maps the grounded states $x \in \mathcal{X}$ to an abstract state $z \in \mathcal{Z}$, where $\mathcal{Z}$ is considered an abstract set of $\mathcal{X}$; that is, we want to recover $S$ through $Z$ by learning an abstraction model. 
For our demonstration, we incorporate our discrete learning framework into the method of learning continuous Markov abstraction \citep{allen2021learning}, which learns an abstraction mapping $\phi$ from the observation space to abstract state space while preserving the Markov property of the original environment.

\begin{figure}[h!]
\begin{centering}
\subfloat{\centering{}\includegraphics[width=1.0\linewidth]{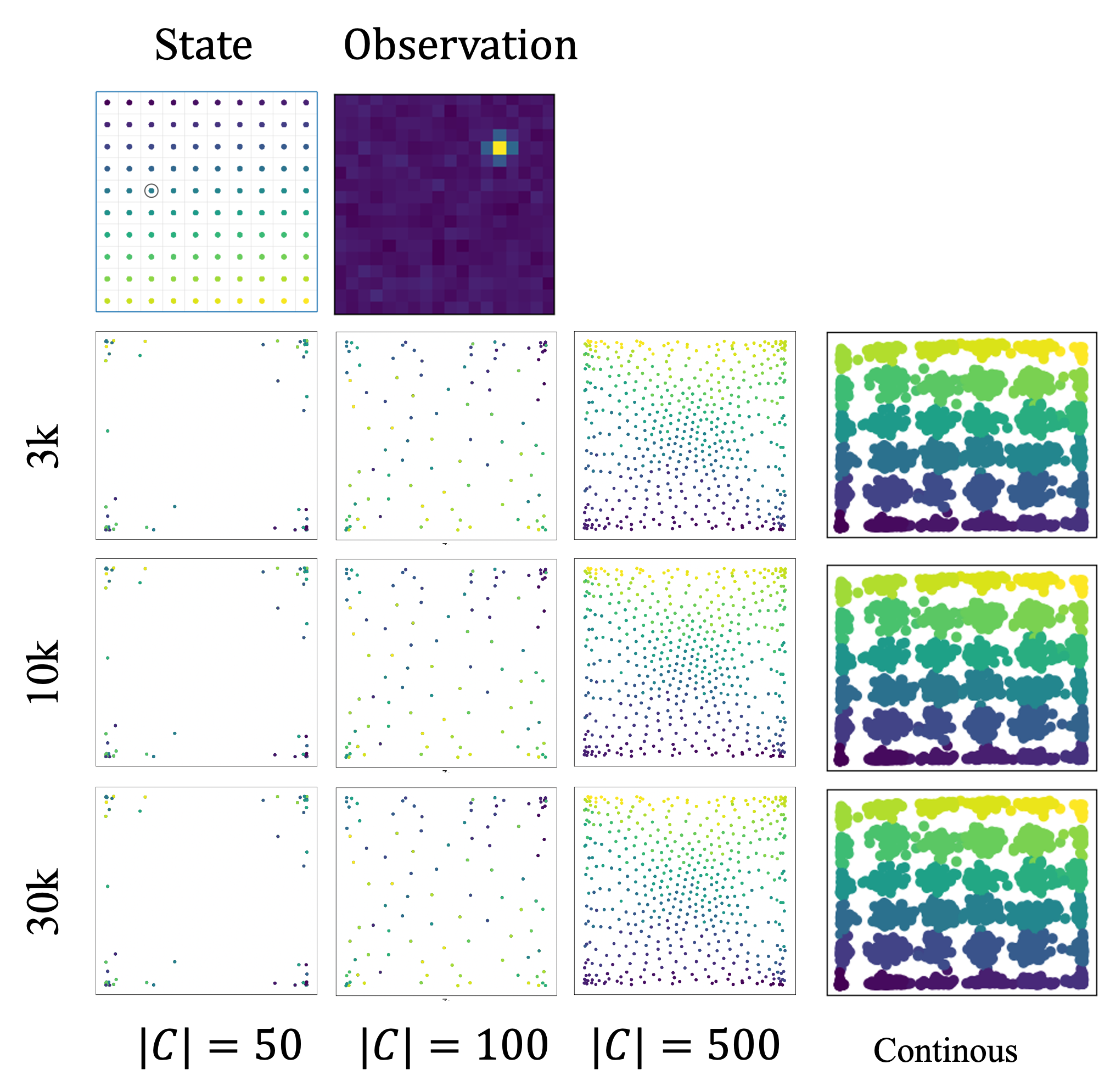}}
\par\end{centering}
\caption{\textbf{(Top row)} The visualization true underlying state on the left, the observations which are noisy version of true state in the middle and the task of RL agent i.e., moving from random point to goal (red point) in the right. \textbf{(From second row)} The visualization of the learned latent codebook; from second top to bottom using $3k$, $10k$, and $30k$ samples for training; from left to right: discrete representation with the codebook sizes of 50, 100, 500 and continuous representation from the algorithm of \citep{allen2021learning}.}
\label{fig:gridworld}
\end{figure}

\begin{figure*}[h!]
\begin{centering}
\subfloat{\centering{}\includegraphics[width=1.0\linewidth]{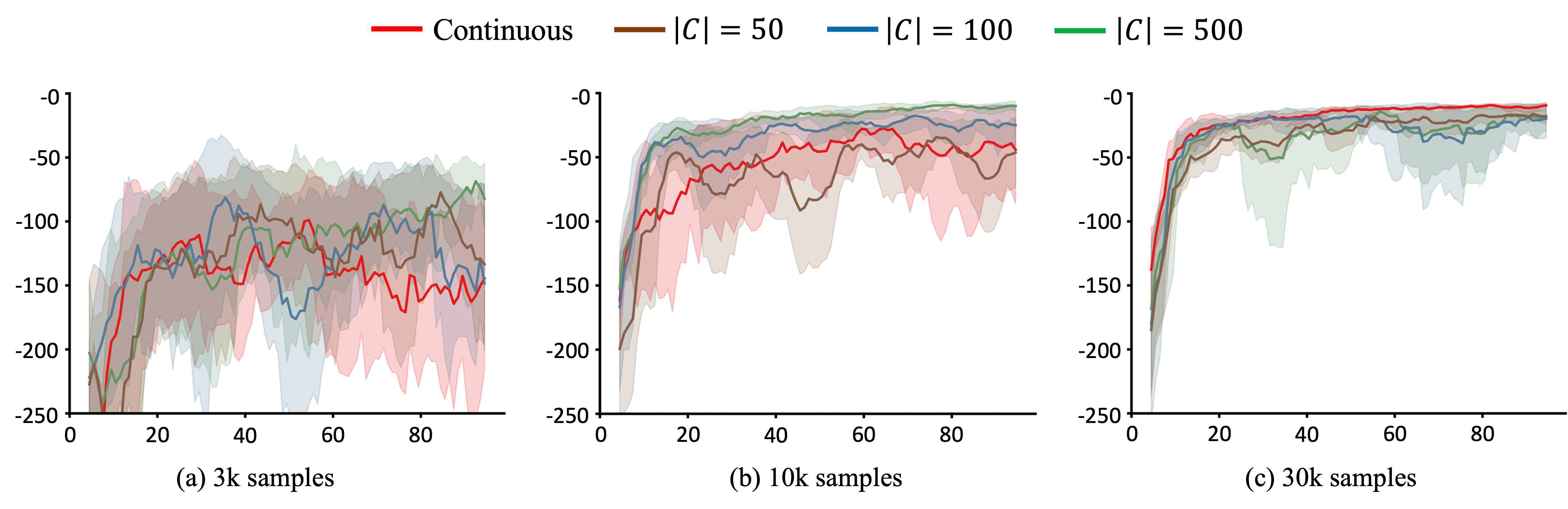}}
\par\end{centering}
\caption{Results on $10\times10$ gridworld environment, tested with different codebook sizes and training samples.}
\label{fig:maze}
\end{figure*}


\textbf{Discrete Representation for State Abstraction.} 
Since there is no reconstruction in an RL problem, we need to define the task $f_y$. 
Given image observation as inputs, the encoder $f_e$ is defined as a feature extractor network, with the output $z$ from the encoder quantized to the nearest codeword through the quantization operator $Q_C$ as in Eq (\ref{eq:reconstruct_form_continuous}), which together form the discrete abstraction model $\phi:=\bar{f}_e=Q_C \circ f_e$. 
The task is to learn the Markov abstraction state through the weighted sum of the Inverse loss and Ratio loss defined in \citep{allen2021learning} as follows.
\begin{align}
\mathcal{L}_{\text{inv}} &= -\sum_i \log f (a=a_i \mid \phi(x), \phi(x'))\\ 
\mathcal L_{\text{ratio}} &= -\sum_i \log(y=y_i | \phi(x), \phi(\tilde{x}))
\end{align}
The inverse loss trains a network to predict the actions taken $a_i$, given the current and the next abstraction $\phi(x)$, $\phi(x')$, and the ratio loss trains another network to predict whether two arbitrary abstractions $\phi(x), \phi(\tilde{x})$ are from two consecutive grounded states, with $y_i$ the binary labels denote if the two states are consecutive. 
Thus, unlike in the reconstruction tasks, the decoders $f_d$ here are defined as the action prediction network in the inverse model and the binary classifier of the next states for estimating density ratios. The labels corresponding to $f_y(x)$ in (\ref{eq:reconstruct_form_continuous}) are the binary labels $y_i$, and the actions $a_i$.

\paragraph{Task.} We evaluate our framework using the offline setting in Visual Gridworld environment from \citep{allen2021learning}. The input image observation of the Gridworld encapsulates the underlying states (Figure~\ref{fig:gridworld} (top row)). An abstraction model is initially trained on offline experiences collected through random exploration without any reward signals. Once trained, the model is frozen, and a DQN agent is trained on the learned latent space using reward signals to navigate from a random starting point to the goal (red point). We compare the agent's performance when using an abstraction model learned through the discrete representation framework versus its continuous counterpart.

\paragraph{Stage-1.} Figure~\ref{fig:gridworld} (starting from the second row) illustrates the representations learned by using discrete-state encoding across different codebook sizes and training sample counts,, compared to the continuous-state representation of the abstraction model (where the continuous representation is considered as a discrete one with a very large codebook) with different number of training sample and codebook sizes. Our experiments show that when the latent codebook size exceeds the number of underlying discrete states, the model successfully reconstructs a meaningful latent structure, reducing noise compared to continuous representations. Even when the environment's state space is larger than the available codebook, the model still learns reasonable representations by organizing similar states into neighboring regions in the latent space.

\paragraph{Stage-2.} Figure \ref{fig:maze} presents the performance of the DQN agent using the state representation learned from stage-1 to evaluate the quality of abstraction model. Comparing discrete-state and continuous-state representations, we observe that with a limited number of samples (e.g., $3k$ samples in \ref{fig:maze}.a and $10k$ samples in Figure \ref{fig:maze}.b), the discrete-state representation outperforms the continuous-state representation. However, with sufficient training data, the continuous-state representation ultimately achieves the best performance. This highlights the trade-off between representation capacity and sample complexity. 

When comparing discrete representations, as shown in Figure \ref{fig:maze}.b with $10k$ samples, it is evident that performance improves as the codebook size increases, given a sufficient number of samples.

\subsection{Domain Generalization}
\label{sec:DG}
\paragraph{Setting.} Given $E$ training source domains $\{\mathbb{P}^{e}(x,y)\}_{e=1}^E$. The objective of DG is to exploit the `commonalities' present in the training domains to improve generalization to any unseen test domain.

\paragraph{Domain Representation Alignment.} Various approaches have been proposed to address the DG problem \citep{long2017conditional, ganin2016domain, li2018domain, gong2016domain, li2018deep, tachet2020domain, arjovsky2020irm, krueger2021out,mitrovic2020representation, wang2022out, yao2022pcl}, in the experiment, we focus in representation alignment-based approach which aim to learn a representation function $f_e$ for data $X$ such that $f_e(X)$ is invariant or consistent across different domains. Two representative work on this approach are DANN \citep{ganin2016domain} which have focused on learning such domain-invariant representations by reducing the divergence between latent marginal distributions $\mathbb{E}[f_e(X) | E]$ where $E$ represents a domain variable and CDANN \citep{li2018domain_b} which aligns the conditional distributions $\mathbb{E}[f_e(X) | Y=y, E]$ across domains. 
 
Let encoder $f_e$, classifier $f_d$, and domain discriminator $D$, the objective for DANN and CDANN can be formulated as follows:
\begin{equation}
\min_{f_e,f_d, D}  \sum_{e=1}^E \mathcal{L}_W(\mathbb{P}^{e}) + \mathcal{L}_D(\mathbb{P}^{e})
\label{eq:dann}
\end{equation}
where \begin{itemize}
    \item $ \mathcal{L}_W(\mathbb{P}^{e})=\mathbb{E}_{(x,y)\sim\mathbb{P}^{e}} \left[  \ell_y \left ( f_d\left (f_e(x)\right), y\right )\right ]$
     \item $ \mathcal{L}_D(\mathbb{P}^{e})=\mathbb{E}_{(x,y)\sim\mathbb{P}^{e}} \left[  \ell_y\left ( D\left(\mathcal{R}\left(f_e(x)\right) ,\phi(x)\right), e\right )\right ]$
\end{itemize}
with $\ell_y$ is cross-entropy-loss; $\mathcal{R}$ is the Gradient Reversal Layer (GRL) as introduced by \citet{ganin2016domain}; $\phi(x)=f_y(x)=y$ for CDANN and \textit{empty} for DANN.

In the above objective, the first term  $\mathcal{L}_W$ is the standard classification loss, while the second term $\mathcal{L}_D$ is the \textit{representation-alignment loss}. Specifically, in the second term, the domain discriminator $D$ aims to accurately predict the domain label ``$e$" based on the combined feature $(f_e(x), m)$. However, with the incorporation of GRL, the gradient that backpropagates from the loss function to the encoder $f_e$ is reversed in the opposite direction. Concurrently, this loss enforces that the representation function $f_e$ transforms the input $x$ into a latent representation $z = f_e(x)$ in such a way that $D$ is unable to determine the domain ``$e$" of $z$, resulting in a domain-invariant representation.

However, achieving true invariance is both challenging and potentially too restrictive. In real-world scenarios, where feature distributions are often multimodal, even feature distributions inner each class. Therefore, directly aligning entire feature distributions or class-conditional distributions across domains can be particularly difficult for adversarial networks. Recent studies \citep{goodfellow2014generative, che2016mode} have shown that adversarial networks are prone to failure by aligning only a subset of the distribution's components (modes), leaving other parts mismatched. To address this problem, our idea here is using set of discrete representation to capture modalities of training-domain distribution, then aligning representation in modality-wise manner. In the following, we demonstrate how to achieve fine-grained modalities of the training-domain distribution by leveraging the multi-level discrete representation property established in Lemma~\ref{cor:distortion_data}.

\paragraph{Fine-grained Representation Alignment.}
\begin{figure}[h!]
\begin{centering}
\subfloat{\centering{}\includegraphics[width=0.7\linewidth]{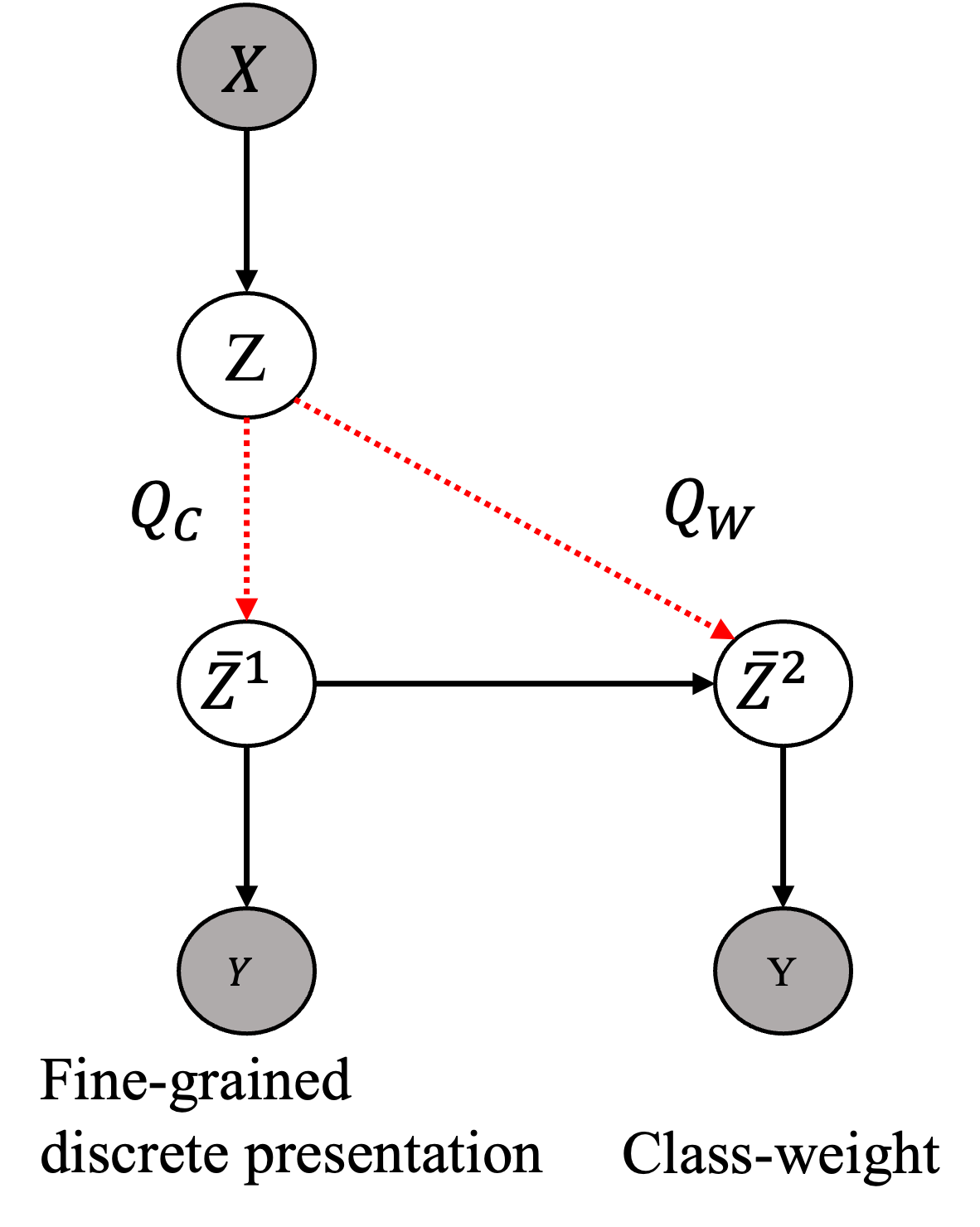}}
\par\end{centering}
\caption{Fine-Grained Discrete Representation: $\bar{Z}^2$ corresponds to  class-level weights, capturing information at the class level, while $\bar{Z}^1$ corresponds to fine-grained discrete representations, characterizing data at a more detailed, modality-specific level.}
\label{fig:DG}
\end{figure}
First, let's reformulate the $K$-class classification problem within our discrete framework. We consider the classifier on top of the encoder as a linear layer represented by the weight matrix $f_d = W = [w_k]_{k=1}^K$, where $w_k$ is the vector corresponding to $k^{th}$ class. The prediction via softmax activation is given by:
\begin{equation*}
    P(y = k \mid x) = \frac{\exp\left(f_e(x)^\top w_k\right)}{\sum_{j=1}^{K} \exp\left(f_e(x)^\top w_j\right)}
\end{equation*}
Therefore, given that $f^{*}_e$ and $W^*$ are the optimal solutions for the classification problem, i.e., $f^*_e, W^* = \underset{f_e,W}{\text{argmin}} \sum_{e=1}^E \mathcal{L}_W(\mathbb{P}^{e})$, the classification process for a given sample-label pair $(x, y)$ is essentially performed by mapping the representation $z = f^*_e(x)$ to the closest class-vector $w^*_{k=y}$ using the metric $d_z(z, w_k) = \exp\left(f_e(x)^\top w_k\right)$ and applying the decoder $f_d(w^*_{k=y})=y$ (mapping class-vector $w_y$ to its index). Consequently, we can formulate the $K$-class classification problem as a discrete representation problem\footnote{We demonstrate the equivalence between $k$-class classification problem and the discrete representation learning presented in OP in Eq.~(\ref{eq:single_level}) in Appendix~\ref{sec:kclass}.}, where $W = \{w_1,...,w_K\}$ is the set of discrete representations, capturing the characteristics of the original data, i.e., a high-level clustering where all samples in a cluster share the same class label $f_y(x)$.

Our goal is to further capture the multiple modalities within the feature distribution of each class. To achieve this, we propose learning an additional set of fine-grained discrete representations $C = \left[c_1, \dots, c_M\right]$ with $M > K$, thereby increasing the codebook size to enhance representation capacity (as shown in Lemma~\ref{cor:distortion_data}). 
This process is illustrated in Figure~\ref{fig:DG}, where $z = f_e(x)$ is the learned representation of input $x$, and $z$ is simultaneously quantized using both the codebook $C$, i.e., $\bar{z}^1 = Q_C(z)$, and the codebook $W$, i.e., $\bar{z}^2 = Q_W(z)$. Note that, by OP in Eq~(\ref{eq:single_level}), both $f_d(\bar{z}^1)$ and $f_d(\bar{z}^1)$ should correctly predict label $f_y(x)$ (the first term of Eq~(\ref{eq:single_level})). Consequently, the continuous representation $z$, which are quantized to $\bar{z}^1$, should belong to the same class as $\bar{z}^2$, forming a sub-class representing by $\bar{z}^2$, effectively partitioning each class into several sub-classes, where each sub-class represents a distinct modality of the class's feature distribution.

Since second level discrete representation is the classification, hence we have the final objective as follows:
\begin{equation}
\min_{f_e,f_d, d}  \sum_{e=1}^E \mathcal{L}_W(\mathbb{P}^{e}) + \mathcal{L}_C(\mathbb{P}^{e}) +\mathcal{L}_D(\mathbb{P}^{e})
\end{equation}
where \begin{itemize}
    \item $ \mathcal{L}_W(\mathbb{P}^{e})=\mathbb{E}_{(x,y)\sim\mathbb{P}^{e}} \left[  \ell_y \left ( f_d\left (f_e(x)\right), y\right )\right ]$
    \item $ \mathcal{L}_C(\mathbb{P}^{e})= 
    \begin{bmatrix}
    \mathbb{E}_{(x,y)\sim\mathbb{P}^{e}} \left[  \ell_y \left ( f_d\left( Q_C(f_e(x))\right), y\right )\right ]
 \\
 + \lambda \mathcal{W}_{d_{z}}\left( f_e\#\mathbb{P}^e(x),\mathbb{P}_{c,\pi}\right)
\end{bmatrix}$
     \item $ \mathcal{L}_D(\mathbb{P}^{e})=\mathbb{E}_{(x,y)\sim\mathbb{P}^{e}} \left[  \ell_y\left ( D\left(\mathcal{R}\left(f_e(x)\right) ,\phi(x)\right), e\right )\right ]$
\end{itemize}

Compared to the objective in Eq.~(\ref{eq:dann}), an additional term, $\mathcal{L}_C$, is introduced to learn a set of discrete representations $C = \{c_1, \dots, c_M\}$, where $\phi(x) = \underset{m}{\text{argmin}} d_z\left(f_e(x), c_m\right)$ is the index of the discrete representation $c_m$ to which sample $x$ is assigned. This leads to a key distinction in the alignment strategy: : DANN aligns the entire domain representation, CDANN aligns class-conditional representations, while this approach utilizes fine-grained discrete representations $C = \{c_1, \dots, c_M\}$, rather than class-level representations $W = \{w_1, \dots, w_K\}$ for conditional alignment, we refer to this method as FDANN.


\begin{table}[h!]
\caption{Classification Accuracy on \textbf{VLCS} using ResNet50. Results are reported over 5 random seeds.}
\begin{centering}
\resizebox{1.0\columnwidth}{!}{ %
\begin{tabular}{lccccc}
\toprule
\textbf{Algorithm}  & \textbf{C} & \textbf{L} & \textbf{S} & \textbf{V} & \textbf{Avg}  \\
\midrule
ERM& {98.8} $\pm$ 0.1 & 63.3 $\pm$ 0.3 & 75.3 $\pm$ 0.5 & 79.2 $\pm$ 0.6 & {79.1}\\
DANN& \textbf{99.2} $\pm$ 0.1 & 63.0 $\pm$ 0.8 & 75.3 $\pm$ 1.8 & 79.3 $\pm$ 0.5 & {79.2}\\
CDANN& {99.1} $\pm$ 0.1 & 63.3 $\pm$ 0.7 & 75.1 $\pm$ 0.7 & 80.1 $\pm$ 0.2 & {79.3}\\
FDANN & {98.9} $\pm$ 0.2 &  \textbf{63.7} $\pm$ 0.3 & 75.6 $\pm$ 0.4 &  \textbf{79.4} $\pm$ 0.8 &  \textbf{79.4} \\
\bottomrule
\end{tabular}}
\par\end{centering}
\label{tab:VLCS}
\end{table}


\begin{table}[h!]
\caption{Classification Accuracy on \textbf{PACS} using ResNet50. Results are reported over 5 random seeds.}
\begin{centering}
\resizebox{1.0\columnwidth}{!}{ %
\begin{tabular}{lccccc}
\toprule
\textbf{Algorithm}  & \textbf{A} & \textbf{C} & \textbf{P} & \textbf{S} & \textbf{Avg}  \\
\midrule
ERM& 89.3 $\pm$ 0.2 & 83.4 $\pm$ 0.6 & 97.3 $\pm$ 0.3 & 82.5 $\pm$ 0.5 & 88.1\\
DANN&  \textbf{90.7} $\pm$ 1.2 & 82.2 $\pm$ 0.4 & 97.3 $\pm$ 0.1 & 81.6 $\pm$ 0.4 & 87.9\\
CDANN & 90.5 $\pm$ 0.3 & 82.4 $\pm$ 1.0 & 97.6 $\pm$ 0.1 & 80.4 $\pm$ 0.3 & 87.7\\
FDANN & 90.5 $\pm$ 0.5 &  \textbf{83.4} $\pm$ 0.2 &  \textbf{97.8} $\pm$ 0.1 &  \textbf{83.2} $\pm$ 0.2 &  \textbf{88.7} \\
\bottomrule
\end{tabular}}
\par\end{centering}
\label{tab:PACS}
\end{table}

\begin{table}[h!]
\caption{Classification Accuracy on \textbf{OfficeHome} using ResNet50. Results are reported over 5 random seeds.}
\begin{centering}
\resizebox{1.0\columnwidth}{!}{ %
\begin{tabular}{lccccc}
\toprule
\textbf{Algorithm}  & \textbf{A} & \textbf{C} & \textbf{P} & \textbf{R} & \textbf{Avg}  \\
\midrule
ERM & 66.1 $\pm$ 0.4 & 57.7 $\pm$ 0.4 & 78.4 $\pm$0.1 & 80.2 $\pm$ 0.2& 70.6\\
DANN& 67.2 $\pm$ 0.1 & 56.2 $\pm$ 0.1 & 78.6 $\pm$0.2 & 80.0 $\pm$ 0.5& 70.5\\
CDANN& 66.8 $\pm$ 0.4 & 56.4 $\pm$ 0.8 & 78.4 $\pm$0.5 & 80.1 $\pm$ 0.2& 70.4\\
FDANN & \textbf{69.1} $\pm$ 0.6 & \textbf{58.4} $\pm$ 0.8 &  \textbf{79.5} $\pm$ 0.2 &  \textbf{81.4} $\pm$ 0.3 & \textbf{72.1}\\
\bottomrule
\end{tabular}}
\par\end{centering}
\label{tab:OfficeHome}
\end{table}

\paragraph{Benchmark.} Table \ref{tab:VLCS}, Table \ref{tab:PACS} and  Table \ref{tab:OfficeHome} present the results of our experiments, comparing FDANN with baseline ERM \citep{gulrajani2020search}, DANN and CDANN across three datasets VLCS, PACS and OfficeHome from the DomainBed benchmark \citep{gulrajani2020search} (the detailed settings are provided in Appendix~\ref{apd:expsetting}). Our method consistently outperforms the baseline approaches on all datasets, demonstrating the effectiveness of aligning representations based on the modalities captured by the set of fine-grained discrete representations.

\begin{table}[h!]
\caption{Classification Accuracy on \textbf{PACS} with different codebook size  using ResNet50. Results are reported over 5 random seeds.}
\begin{centering}
\resizebox{1.0\columnwidth}{!}{ %
\begin{tabular}{lccccc}
\toprule
M  & \textbf{A} & \textbf{C} & \textbf{P} & \textbf{S} & \textbf{Avg}  \\
\midrule
CDANN & 90.5 $\pm$ 0.3 & 82.4 $\pm$ 1.0 & 97.6 $\pm$ 0.1 & 80.4 $\pm$ 0.3 & 87.7\\
$K\times 4$ & 90.2 $\pm$ 0.3 & 83.2 $\pm$ 0.7 & 97.9 $\pm$ 0.2 & 82.3 $\pm$ 1.5 & 88.2 \\

$K\times 8$ & 90.5 $\pm$ 0.8 & 83.8 $\pm$ 0.6 & 97.6 $\pm$ 0.3 &  82.1 $\pm$ 1.8 &   \textbf{88.7}\\

$K\times 16$ & 90.5 $\pm$ 0.5 & 83.4 $\pm$ 0.2 & 97.8 $\pm$ 0.1 & 83.2 $\pm$ 0.2 &  \textbf{88.7} \\
$K\times 32$ & 90.2 $\pm$ 0.5&  83.8 $\pm$ 0.8 & 97.3 $\pm$ 0.4 & 82.0 $\pm$ 1.2 & 88.4\\
\bottomrule
\end{tabular}}
\par\end{centering}
\label{tab:PACS_prototype}
\end{table}

\paragraph{Effect of Codebook Size.} We conducted an additional ablation study to further investigate the effect of codebook size on performance. It is important to note that CDANN can be considered a special case of our FDANN when $M = K\times 1$. As shown in Table \ref{tab:PACS_prototype}, performance generally improves as the codebook size increases. However, when $K$ becomes excessively large (i.e., $M = K*32$), a decline is observed, likely due to trade-offs in the optimization process.

\section{Limitations and Conclusion}

In this work, we introduce a generalized framework for learning deep discrete representations from a task-driven perspective. We provide a theoretical analysis of the learned discrete representations, emphasizing the trade-off between optimal task accuracy and sample complexity. Specifically, reducing the codebook size lowers the achievable accuracy but requires fewer samples to generalize effectively. Conversely, increasing the codebook size enhances accuracy but demands a larger sample size. We translate these theoretical findings into practical insights for optimizing discrete latents in specific downstream tasks. Finally, we perform extensive experiments to validate our analyses. However, our theoretical framework relies on the use of Wasserstein distances to align latent spaces. The computational complexity of calculating the Wasserstein distance is well-documented and can introduce significant overhead to the learning process. In future work, we aim to explore alternative methods to develop a more efficient learning framework.

\subsubsection*{Acknowledgements}
We sincerely thank our colleagues Bang Giang Le, Viet Cuong Ta, Trung Le, and Dinh Phung for their valuable discussions and support throughout the paper writing and experiment implementation.

\bibliography{references}

\begin{thebibliography}{58}
\providecommand{\natexlab}[1]{#1}
\providecommand{\url}[1]{\texttt{#1}}
\expandafter\ifx\csname urlstyle\endcsname\relax
  \providecommand{\doi}[1]{doi: #1}\else
  \providecommand{\doi}{doi: \begingroup \urlstyle{rm}\Url}\fi

\bibitem[Agarwal et~al.(2019)Agarwal, Jiang, Kakade, and Sun]{agarwal2019reinforcement}
Alekh Agarwal, Nan Jiang, Sham~M Kakade, and Wen Sun.
\newblock Reinforcement learning: Theory and algorithms.
\newblock \emph{CS Dept., UW Seattle, Seattle, WA, USA, Tech. Rep}, 32:\penalty0 96, 2019.

\bibitem[Allen et~al.(2021)Allen, Parikh, Gottesman, and Konidaris]{allen2021learning}
Cameron Allen, Neev Parikh, Omer Gottesman, and George Konidaris.
\newblock Learning markov state abstractions for deep reinforcement learning.
\newblock \emph{Advances in Neural Information Processing Systems}, 34:\penalty0 8229--8241, 2021.

\bibitem[Arjovsky et~al.(2020)Arjovsky, Bottou, Gulrajani, and Lopez-Paz]{arjovsky2020irm}
Martin Arjovsky, Léon Bottou, Ishaan Gulrajani, and David Lopez-Paz.
\newblock Invariant risk minimization, 2020.
\newblock URL \url{https://arxiv.org/abs/1907.02893}.

\bibitem[Bachem et~al.(2017)Bachem, Lucic, Hassani, and Krause]{bachem2017uniform}
Olivier Bachem, Mario Lucic, S~Hamed Hassani, and Andreas Krause.
\newblock Uniform deviation bounds for k-means clustering.
\newblock In \emph{International conference on machine learning}, pp.\  283--291. PMLR, 2017.

\bibitem[Beery et~al.(2018)Beery, Van~Horn, and Perona]{beery2018recognition}
Sara Beery, Grant Van~Horn, and Pietro Perona.
\newblock Recognition in terra incognita.
\newblock In \emph{Proceedings of the European conference on computer vision (ECCV)}, pp.\  456--473, 2018.

\bibitem[Cha et~al.(2021)Cha, Chun, Lee, Cho, Park, Lee, and Park]{cha2021swad}
Junbum Cha, Sanghyuk Chun, Kyungjae Lee, Han-Cheol Cho, Seunghyun Park, Yunsung Lee, and Sungrae Park.
\newblock Swad: Domain generalization by seeking flat minima.
\newblock \emph{Advances in Neural Information Processing Systems}, 34:\penalty0 22405--22418, 2021.

\bibitem[Che et~al.(2016)Che, Li, Jacob, Bengio, and Li]{che2016mode}
Tong Che, Yanran Li, Athul~Paul Jacob, Yoshua Bengio, and Wenjie Li.
\newblock Mode regularized generative adversarial networks.
\newblock \emph{arXiv preprint arXiv:1612.02136}, 2016.

\bibitem[Chen et~al.(2020)Chen, Kornblith, Norouzi, and Hinton]{chen2020simple}
Ting Chen, Simon Kornblith, Mohammad Norouzi, and Geoffrey Hinton.
\newblock A simple framework for contrastive learning of visual representations.
\newblock In \emph{International conference on machine learning}, pp.\  1597--1607. PMLR, 2020.

\bibitem[Chen \& He(2021)Chen and He]{chen2021exploring}
Xinlei Chen and Kaiming He.
\newblock Exploring simple siamese representation learning.
\newblock In \emph{Proceedings of the IEEE/CVF Conference on Computer Vision and Pattern Recognition}, pp.\  15750--15758, 2021.

\bibitem[Dhariwal et~al.(2020)Dhariwal, Jun, Payne, Kim, Radford, and Sutskever]{dhariwal2020jukebox}
Prafulla Dhariwal, Heewoo Jun, Christine Payne, Jong~Wook Kim, Alec Radford, and Ilya Sutskever.
\newblock Jukebox: A generative model for music.
\newblock \emph{arXiv preprint arXiv:2005.00341}, 2020.

\bibitem[Esser et~al.(2021)Esser, Rombach, and Ommer]{esser2021taming}
Patrick Esser, Robin Rombach, and Bjorn Ommer.
\newblock Taming transformers for high-resolution image synthesis.
\newblock In \emph{Proceedings of the IEEE/CVF Conference on Computer Vision and Pattern Recognition}, pp.\  12873--12883, 2021.

\bibitem[Ganin et~al.(2016)Ganin, Ustinova, Ajakan, Germain, Larochelle, Laviolette, Marchand, and Lempitsky]{ganin2016domain}
Yaroslav Ganin, Evgeniya Ustinova, Hana Ajakan, Pascal Germain, Hugo Larochelle, Fran{\c{c}}ois Laviolette, Mario Marchand, and Victor Lempitsky.
\newblock Domain-adversarial training of neural networks.
\newblock \emph{The Journal of Machine Learning Research}, 17\penalty0 (1):\penalty0 2096--2030, 2016.

\bibitem[Gong et~al.(2016)Gong, Zhang, Liu, Tao, Glymour, and Sch{\"o}lkopf]{gong2016domain}
Mingming Gong, Kun Zhang, Tongliang Liu, Dacheng Tao, Clark Glymour, and Bernhard Sch{\"o}lkopf.
\newblock Domain adaptation with conditional transferable components.
\newblock In \emph{International conference on machine learning}, pp.\  2839--2848. PMLR, 2016.

\bibitem[Goodfellow et~al.(2014)Goodfellow, Pouget-Abadie, Mirza, Xu, Warde-Farley, Ozair, Courville, and Bengio]{goodfellow2014generative}
Ian Goodfellow, Jean Pouget-Abadie, Mehdi Mirza, Bing Xu, David Warde-Farley, Sherjil Ozair, Aaron Courville, and Yoshua Bengio.
\newblock Generative adversarial nets.
\newblock \emph{Advances in neural information processing systems}, 27, 2014.

\bibitem[Gulrajani \& Lopez-Paz(2021)Gulrajani and Lopez-Paz]{gulrajani2020search}
Ishaan Gulrajani and David Lopez-Paz.
\newblock In search of lost domain generalization.
\newblock In \emph{International Conference on Learning Representations}, 2021.

\bibitem[Hafner et~al.(2020)Hafner, Lillicrap, Norouzi, and Ba]{hafner2020mastering}
Danijar Hafner, Timothy Lillicrap, Mohammad Norouzi, and Jimmy Ba.
\newblock Mastering atari with discrete world models.
\newblock \emph{arXiv preprint arXiv:2010.02193}, 2020.

\bibitem[Hsu et~al.(2012)Hsu, Kakade, and Zhang]{hsu2012spectral}
Daniel Hsu, Sham~M Kakade, and Tong Zhang.
\newblock A spectral algorithm for learning hidden markov models.
\newblock \emph{Journal of Computer and System Sciences}, 78\penalty0 (5):\penalty0 1460--1480, 2012.

\bibitem[Hu et~al.(2017)Hu, Miyato, Tokui, Matsumoto, and Sugiyama]{hu2017learning}
Weihua Hu, Takeru Miyato, Seiya Tokui, Eiichi Matsumoto, and Masashi Sugiyama.
\newblock Learning discrete representations via information maximizing self-augmented training.
\newblock In \emph{International conference on machine learning}, pp.\  1558--1567. PMLR, 2017.

\bibitem[Islam et~al.(2022)Islam, Zang, Tomar, Didolkar, Islam, Arnob, Iqbal, Li, Goyal, Heess, et~al.]{islam2022representation}
Riashat Islam, Hongyu Zang, Manan Tomar, Aniket Didolkar, Md~Mofijul Islam, Samin~Yeasar Arnob, Tariq Iqbal, Xin Li, Anirudh Goyal, Nicolas Heess, et~al.
\newblock Representation learning in deep rl via discrete information bottleneck.
\newblock \emph{arXiv preprint arXiv:2212.13835}, 2022.

\bibitem[Kalchbrenner et~al.(2017)Kalchbrenner, Oord, Simonyan, Danihelka, Vinyals, Graves, and Kavukcuoglu]{kalchbrenner2017video}
Nal Kalchbrenner, A{\"a}ron Oord, Karen Simonyan, Ivo Danihelka, Oriol Vinyals, Alex Graves, and Koray Kavukcuoglu.
\newblock Video pixel networks.
\newblock In \emph{International Conference on Machine Learning}, pp.\  1771--1779. PMLR, 2017.

\bibitem[Kantorovich(2006)]{kantorovich2006problem}
LK~Kantorovich.
\newblock On a problem of monge.
\newblock \emph{Journal of Mathematical Sciences}, 133\penalty0 (4), 2006.

\bibitem[Kim et~al.(2022)Kim, Song, Lee, Kim, Seo, Lee, Kim, Lee, and Bae]{kim2022verse}
Taehoon Kim, Gwangmo Song, Sihaeng Lee, Sangyun Kim, Yewon Seo, Soonyoung Lee, Seung~Hwan Kim, Honglak Lee, and Kyunghoon Bae.
\newblock L-verse: Bidirectional generation between image and text.
\newblock In \emph{Proceedings of the IEEE/CVF Conference on Computer Vision and Pattern Recognition}, pp.\  16526--16536, 2022.

\bibitem[Kingma \& Welling(2013)Kingma and Welling]{kingma2013auto}
Diederik~P Kingma and Max Welling.
\newblock Auto-encoding variational bayes.
\newblock \emph{arXiv preprint arXiv:1312.6114}, 2013.

\bibitem[Kingma et~al.(2016)Kingma, Salimans, Jozefowicz, Chen, Sutskever, and Welling]{kingma2016improved}
Durk~P Kingma, Tim Salimans, Rafal Jozefowicz, Xi~Chen, Ilya Sutskever, and Max Welling.
\newblock Improved variational inference with inverse autoregressive flow.
\newblock \emph{Advances in neural information processing systems}, 29, 2016.

\bibitem[Krueger et~al.(2021)Krueger, Caballero, Jacobsen, Zhang, Binas, Zhang, Le~Priol, and Courville]{krueger2021out}
David Krueger, Ethan Caballero, Joern-Henrik Jacobsen, Amy Zhang, Jonathan Binas, Dinghuai Zhang, Remi Le~Priol, and Aaron Courville.
\newblock Out-of-distribution generalization via risk extrapolation (rex).
\newblock In \emph{International Conference on Machine Learning}, pp.\  5815--5826. PMLR, 2021.

\bibitem[Li et~al.(2017)Li, Yang, Song, and Hospedales]{li2017deeper}
Da~Li, Yongxin Yang, Yi-Zhe Song, and Timothy~M Hospedales.
\newblock Deeper, broader and artier domain generalization.
\newblock In \emph{Proceedings of the IEEE international conference on computer vision}, pp.\  5542--5550, 2017.

\bibitem[Li et~al.(2018{\natexlab{a}})Li, Pan, Wang, and Kot]{li2018domain}
Haoliang Li, Sinno~Jialin Pan, Shiqi Wang, and Alex~C Kot.
\newblock Domain generalization with adversarial feature learning.
\newblock In \emph{Proceedings of the IEEE conference on computer vision and pattern recognition}, pp.\  5400--5409, 2018{\natexlab{a}}.

\bibitem[Li et~al.(2018{\natexlab{b}})Li, Gong, Tian, Liu, and Tao]{li2018domain_b}
Ya~Li, Mingming Gong, Xinmei Tian, Tongliang Liu, and Dacheng Tao.
\newblock Domain generalization via conditional invariant representations.
\newblock In \emph{Proceedings of the AAAI conference on artificial intelligence}, volume~32, 2018{\natexlab{b}}.

\bibitem[Li et~al.(2018{\natexlab{c}})Li, Tian, Gong, Liu, Liu, Zhang, and Tao]{li2018deep}
Ya~Li, Xinmei Tian, Mingming Gong, Yajing Liu, Tongliang Liu, Kun Zhang, and Dacheng Tao.
\newblock Deep domain generalization via conditional invariant adversarial networks.
\newblock In \emph{Proceedings of the European conference on computer vision (ECCV)}, pp.\  624--639, 2018{\natexlab{c}}.

\bibitem[Lin et~al.(2023)Lin, Chen, and Wang]{lin2023frequency}
Ci-Siang Lin, Min-Hung Chen, and Yu-Chiang~Frank Wang.
\newblock Frequency-aware self-supervised long-tailed learning.
\newblock In \emph{Proceedings of the IEEE/CVF International Conference on Computer Vision}, pp.\  963--972, 2023.

\bibitem[Liu et~al.(2022)Liu, Lamb, Ji, Notsawo, Mozer, Bengio, and Kawaguchi]{liu2022adaptive}
Dianbo Liu, Alex Lamb, Xu~Ji, Pascal Notsawo, Mike Mozer, Yoshua Bengio, and Kenji Kawaguchi.
\newblock Adaptive discrete communication bottlenecks with dynamic vector quantization.
\newblock \emph{arXiv preprint arXiv:2202.01334}, 2022.

\bibitem[Long et~al.(2017)Long, Cao, Wang, and Jordan]{long2017conditional}
Mingsheng Long, Zhangjie Cao, Jianmin Wang, and Michael~I Jordan.
\newblock Conditional adversarial domain adaptation.
\newblock \emph{arXiv preprint arXiv:1705.10667}, 2017.

\bibitem[Ming et~al.(2022)Ming, Sun, Dia, and Li]{ming2022exploit}
Yifei Ming, Yiyou Sun, Ousmane Dia, and Yixuan Li.
\newblock How to exploit hyperspherical embeddings for out-of-distribution detection?
\newblock \emph{arXiv preprint arXiv:2203.04450}, 2022.

\bibitem[Mitrovic et~al.(2020)Mitrovic, McWilliams, Walker, Buesing, and Blundell]{mitrovic2020representation}
Jovana Mitrovic, Brian McWilliams, Jacob Walker, Lars Buesing, and Charles Blundell.
\newblock Representation learning via invariant causal mechanisms.
\newblock \emph{arXiv preprint arXiv:2010.07922}, 2020.

\bibitem[Oord et~al.(2016)Oord, Dieleman, Zen, Simonyan, Vinyals, Graves, Kalchbrenner, Senior, and Kavukcuoglu]{oord2016wavenet}
Aaron van~den Oord, Sander Dieleman, Heiga Zen, Karen Simonyan, Oriol Vinyals, Alex Graves, Nal Kalchbrenner, Andrew Senior, and Koray Kavukcuoglu.
\newblock Wavenet: A generative model for raw audio.
\newblock \emph{arXiv preprint arXiv:1609.03499}, 2016.

\bibitem[Ozair et~al.(2021)Ozair, Li, Razavi, Antonoglou, Van Den~Oord, and Vinyals]{ozair2021vector}
Sherjil Ozair, Yazhe Li, Ali Razavi, Ioannis Antonoglou, Aaron Van Den~Oord, and Oriol Vinyals.
\newblock Vector quantized models for planning.
\newblock In \emph{international conference on machine learning}, pp.\  8302--8313. PMLR, 2021.

\bibitem[Pathak et~al.(2016)Pathak, Krahenbuhl, Donahue, Darrell, and Efros]{pathak2016context}
Deepak Pathak, Philipp Krahenbuhl, Jeff Donahue, Trevor Darrell, and Alexei~A Efros.
\newblock Context encoders: Feature learning by inpainting.
\newblock In \emph{Proceedings of the IEEE conference on computer vision and pattern recognition}, pp.\  2536--2544, 2016.

\bibitem[Peng et~al.(2019)Peng, Bai, Xia, Huang, Saenko, and Wang]{peng2019moment}
Xingchao Peng, Qinxun Bai, Xide Xia, Zijun Huang, Kate Saenko, and Bo~Wang.
\newblock Moment matching for multi-source domain adaptation.
\newblock In \emph{Proceedings of the IEEE/CVF international conference on computer vision}, pp.\  1406--1415, 2019.

\bibitem[Ramesh et~al.(2021)Ramesh, Pavlov, Goh, Gray, Voss, Radford, Chen, and Sutskever]{ramesh2021zero}
Aditya Ramesh, Mikhail Pavlov, Gabriel Goh, Scott Gray, Chelsea Voss, Alec Radford, Mark Chen, and Ilya Sutskever.
\newblock Zero-shot text-to-image generation.
\newblock In \emph{International conference on machine learning}, pp.\  8821--8831. Pmlr, 2021.

\bibitem[Razavi et~al.(2019)Razavi, Van~den Oord, and Vinyals]{razavi2019generating}
Ali Razavi, Aaron Van~den Oord, and Oriol Vinyals.
\newblock Generating diverse high-fidelity images with vq-vae-2.
\newblock \emph{Advances in neural information processing systems}, 32, 2019.

\bibitem[Reed et~al.(2017)Reed, Oord, Kalchbrenner, Colmenarejo, Wang, Chen, Belov, and Freitas]{reed2017parallel}
Scott Reed, A{\"a}ron Oord, Nal Kalchbrenner, Sergio~G{\'o}mez Colmenarejo, Ziyu Wang, Yutian Chen, Dan Belov, and Nando Freitas.
\newblock Parallel multiscale autoregressive density estimation.
\newblock In \emph{International conference on machine learning}, pp.\  2912--2921. PMLR, 2017.

\bibitem[Rombach et~al.(2022)Rombach, Blattmann, Lorenz, Esser, and Ommer]{rombach2022high}
Robin Rombach, Andreas Blattmann, Dominik Lorenz, Patrick Esser, and Bj{\"o}rn Ommer.
\newblock High-resolution image synthesis with latent diffusion models.
\newblock In \emph{Proceedings of the IEEE/CVF conference on computer vision and pattern recognition}, pp.\  10684--10695, 2022.

\bibitem[Roy et~al.(2018)Roy, Vaswani, Neelakantan, and Parmar]{roy2018theory}
Aurko Roy, Ashish Vaswani, Arvind Neelakantan, and Niki Parmar.
\newblock Theory and experiments on vector quantized autoencoders.
\newblock \emph{arXiv preprint arXiv:1805.11063}, 2018.

\bibitem[Santambrogio(2015)]{santambrogio2015optimal}
Filippo Santambrogio.
\newblock Optimal transport for applied mathematicians.
\newblock \emph{Birk{\"a}user, NY}, 55\penalty0 (58-63):\penalty0 94, 2015.

\bibitem[Tachet~des Combes et~al.(2020)Tachet~des Combes, Zhao, Wang, and Gordon]{tachet2020domain}
Remi Tachet~des Combes, Han Zhao, Yu-Xiang Wang, and Geoffrey~J Gordon.
\newblock Domain adaptation with conditional distribution matching and generalized label shift.
\newblock \emph{Advances in Neural Information Processing Systems}, 33:\penalty0 19276--19289, 2020.

\bibitem[Takida et~al.(2022)Takida, Shibuya, Liao, Lai, Ohmura, Uesaka, Murata, Takahashi, Kumakura, and Mitsufuji]{takida22a}
Yuhta Takida, Takashi Shibuya, Weihsiang Liao, Chieh-Hsin Lai, Junki Ohmura, Toshimitsu Uesaka, Naoki Murata, Shusuke Takahashi, Toshiyuki Kumakura, and Yuki Mitsufuji.
\newblock {SQ}-{VAE}: Variational {B}ayes on discrete representation with self-annealed stochastic quantization.
\newblock In Kamalika Chaudhuri, Stefanie Jegelka, Le~Song, Csaba Szepesvari, Gang Niu, and Sivan Sabato (eds.), \emph{Proceedings of the 39th International Conference on Machine Learning}, volume 162 of \emph{Proceedings of Machine Learning Research}, pp.\  20987--21012. PMLR, 17--23 Jul 2022.

\bibitem[Telgarsky \& Dasgupta(2013)Telgarsky and Dasgupta]{telgarsky2013moment}
Matus~J Telgarsky and Sanjoy Dasgupta.
\newblock Moment-based uniform deviation bounds for $ k $-means and friends.
\newblock \emph{Advances in Neural Information Processing Systems}, 26, 2013.

\bibitem[Torralba \& Efros(2011)Torralba and Efros]{torralba2011unbiased}
Antonio Torralba and Alexei~A Efros.
\newblock Unbiased look at dataset bias.
\newblock In \emph{CVPR 2011}, pp.\  1521--1528. IEEE, 2011.

\bibitem[Van Den~Oord et~al.(2017)Van Den~Oord, Vinyals, et~al.]{van2017neural}
Aaron Van Den~Oord, Oriol Vinyals, et~al.
\newblock Neural discrete representation learning.
\newblock \emph{Advances in neural information processing systems}, 30, 2017.

\bibitem[Venkateswara et~al.(2017)Venkateswara, Eusebio, Chakraborty, and Panchanathan]{venkateswara2017deep}
Hemanth Venkateswara, Jose Eusebio, Shayok Chakraborty, and Sethuraman Panchanathan.
\newblock Deep hashing network for unsupervised domain adaptation.
\newblock In \emph{Proceedings of the IEEE conference on computer vision and pattern recognition}, pp.\  5018--5027, 2017.

\bibitem[Vuong et~al.(2023)Vuong, Le, Zhao, Zheng, Harandi, Cai, and Phung]{vuong2023vector}
Tung-Long Vuong, Trung Le, He~Zhao, Chuanxia Zheng, Mehrtash Harandi, Jianfei Cai, and Dinh Phung.
\newblock Vector quantized wasserstein auto-encoder.
\newblock \emph{arXiv preprint arXiv:2302.05917}, 2023.

\bibitem[Wang et~al.(2022)Wang, Yi, Chen, and Zhu]{wang2022out}
Ruoyu Wang, Mingyang Yi, Zhitang Chen, and Shengyu Zhu.
\newblock Out-of-distribution generalization with causal invariant transformations.
\newblock In \emph{Proceedings of the IEEE/CVF Conference on Computer Vision and Pattern Recognition}, pp.\  375--385, 2022.

\bibitem[Williams et~al.(2020)Williams, Ringer, Ash, MacLeod, Dougherty, and Hughes]{williams2020hierarchical}
Will Williams, Sam Ringer, Tom Ash, David MacLeod, Jamie Dougherty, and John Hughes.
\newblock Hierarchical quantized autoencoders.
\newblock \emph{Advances in Neural Information Processing Systems}, 33:\penalty0 4524--4535, 2020.

\bibitem[Yao et~al.(2022)Yao, Bai, Zhang, Zhang, Sun, Chen, Li, and Yu]{yao2022pcl}
Xufeng Yao, Yang Bai, Xinyun Zhang, Yuechen Zhang, Qi~Sun, Ran Chen, Ruiyu Li, and Bei Yu.
\newblock Pcl: Proxy-based contrastive learning for domain generalization.
\newblock In \emph{Proceedings of the IEEE/CVF Conference on Computer Vision and Pattern Recognition}, pp.\  7097--7107, 2022.

\bibitem[Yarats et~al.(2021)Yarats, Fergus, Lazaric, and Pinto]{yarats2021reinforcement}
Denis Yarats, Rob Fergus, Alessandro Lazaric, and Lerrel Pinto.
\newblock Reinforcement learning with prototypical representations.
\newblock In \emph{International Conference on Machine Learning}, pp.\  11920--11931. PMLR, 2021.

\bibitem[Ye et~al.(2024)Ye, Fan, Song, Zheng, Zhao, Guo, and Chang]{ye2024ptarl}
Hangting Ye, Wei Fan, Xiaozhuang Song, Shun Zheng, He~Zhao, Dandan Guo, and Yi~Chang.
\newblock Ptarl: Prototype-based tabular representation learning via space calibration.
\newblock \emph{arXiv preprint arXiv:2407.05364}, 2024.

\bibitem[Yu et~al.(2021)Yu, Li, Koh, Zhang, Pang, Qin, Ku, Xu, Baldridge, and Wu]{yu2021vector}
Jiahui Yu, Xin Li, Jing~Yu Koh, Han Zhang, Ruoming Pang, James Qin, Alexander Ku, Yuanzhong Xu, Jason Baldridge, and Yonghui Wu.
\newblock Vector-quantized image modeling with improved vqgan.
\newblock In \emph{International Conference on Learning Representations}, 2021.

\bibitem[Zheng et~al.(2022)Zheng, Vuong, Cai, and Phung]{Zheng2022MOvq}
Chuanxia Zheng, Long~Tung Vuong, Jianfei Cai, and Dinh Phung.
\newblock Movq: Modulating quantized vectors for high-fidelity image generation.
\newblock \emph{Advances in Neural Information Processing Systems}, 35, 2022.

\end{thebibliography}
\bibliographystyle{iclr2025_conference}

\onecolumn
\section*{Appendix}

The appendix provides supplementary material and additional details to support the main findings and methods proposed in this paper. It is organized into several sections:

Section 6 provides the proofs for the theoretical results discussed in the main paper. Specifically:
\begin{itemize}
    \item Section \ref{sec:proof-1} presents the proof of Theorem \ref{thm:trainable_single}, which transforms the constrained discrete representation learning framework into a trainable form.
    \item Section \ref{sec:proof-2} provides the proof for the Lemma ~\ref{cor:distortion_data} on the accuracy trade-off.
    \item Section \ref{sec:proof-3} includes the Corollary~\ref{cor:sample_complexity} on sample complexity.
    \item Finally, Section \ref{sec:proof-4} demonstrates the equivalence between the common classification problem and the discrete representation framework.
\end{itemize}

Section \ref{apd:expsetting} outlines the details of the experimental setup, including datasets and hyperparameters. The source code is provided in the supplementary material.
\section{Theoretical development}
 \label{apd:theory}
In this Section, we present all proofs relevant to theory developed
in our paper.

\paragraph{Notations.} We use calligraphic letters (i.e., $\mathcal{X}$) for spaces, upper case letters (i.e. $X$) for random variables, lower case letters (i.e. $x$) for their values and $\mathbb{P}$ for (observed) probability distributions. We consider the task-driven setting where the observation distribution is $\mathbb{P}(x)$ (or $\mathbb{P}_x$) and labelling function $f_y(\cdot)$ is a deterministic mapping from $x$ to the outcomes/labels the task of interest (e.g., a categorical label for a classification problem, a real value/vector for a regression problem).

\textbf{Objective function.}  The OP of discrete representation learning 
framework w.r.t training data $\mathbb{P}_x$ is formulated as the minimizer of following OP:
\begin{align}
\min_{\mathbb{P}_{c,\pi},f_{d}}\min_{\bar{f}_{e}:\bar{f}_{e}\#\mathbb{P}_{x}=\mathbb{P}_{c,\pi}}   \underset{x\sim\mathbb{P}_{x}}{\mathbb{E}}\left[\ell\left(f_{d}\left(\bar{f}_{e}\left(x\right)\right),f_y(x)\right)\right] 
\label{eq:single_level_apd}
\end{align}
Where
\begin{itemize}
    \item $\mathbb{P}_{c,\pi}= \sum_{m=1}^{M}\pi_{m}\delta_{c_{m}}$ is the discrete distriubtion over the codebook $C=\{c_{m}\}_{m=1}^{M}$ where  the category probabilities $\pi \in \Delta_{M} = \{\alpha \ge 0 : \Vert \alpha \Vert_1 = 1 \}$ lie in the $M$-simplex.
    
    \item The discrete representation function $\bar{f}_e=Q_C \circ f_e$ in which the \emph{encoder} $f_{e}: \mathcal{X}\rightarrow \mathcal{Z}$ first map the data examples $x\in \mathcal{X}$ to the latent $z\in \mathcal{Z}$. Following, a quantization $Q_C$ projecting $z$ onto $C: \bar{z} =Q_C(z) = \text{argmin}{}_{c\in C}d_{z}\left(z,c\right)$ is a quantization operator which $d_z$ is a distance on $\mathcal{Z}$. 

    \item The decoder $f_d$ using discrete representation $\bar{z}$ to make prediction and $\ell$ is the loss of interest.
\end{itemize}

\subsection{Proof theorem~\ref{thm:trainable_single} in the main paper}
\label{sec:proof-1}

\begin{lemma}
\label{measure_lemma}
Consider $C, \pi, f_d$, and $f_e$ as a feasible solution of the OP in Eq. (\ref{eq:single_level_apd}). Let us denote $\bar{f}_e(x) = argmin_{c} d_z(f_e(x)),c) = Q_C(x)$, then $\bar{f}_e(x)$ is a Borel measurable function. 
\end{lemma}

\begin{proof}
We denote the set $A_k$ on the latent space as 
\begin{equation*}
A_k = \{z: d_z(z,c_k) < d(z, c_j), \forall j\neq k \} = \{z: Q_C(z) = c_k\}.    
\end{equation*}
$A_k$ is known as  a Voronoi cell w.r.t. the metric $d_z$. If we consider a continuous metric $d_z$, $A_k$ is a measurable set. Given a Borel measurable function $B$, we prove that $\bar{f}_e^{-1}(B)$ is a Borel measurable set on the data space.

Let $B\cap\{c_{1},..,c_{K}\}=\{c_{i_{1}},...,c_{i_{m}}\}$, we prove that $\bar{f}_{e}^{-1}\left(B\right)=\cup_{j=1}^{m}f_{e}^{-1}\left(A_{i_{j}}\right)$. Indeed, take $x \in \bar{f}_{e}^{-1}\left(B\right)$, then $\bar{f}_{e}(x) \in B$, implying that $\bar{f}_{e}(x) = Q_C(x) = c_{i_j}$ for some $j=1,...,m$. This means that $f_e(x) \in A_{i_j}$ for some $j=1,...,m$. Therefore, we reach $\bar{f}_{e}^{-1}\left(B\right) \subset\cup_{j=1}^{m}f_{e}^{-1}\left(A_{i_{j}}\right)$.

We now take $x \in \cup_{j=1}^{m}f_{e}^{-1}\left(A_{i_{j}}\right)$. Then $f_e(x) \in A_{i_j}$ for $j=1,...,m$, hence $\bar{f}_{e}(x) = Q_C(x) = c_{i_j}$ for some $j=1,...,m$. Thus, $\bar{f}_{e}(x) \subset B$ or equivalently $x \in \bar{f}_{e}^{-1}\left(B\right)$, implying $\bar{f}_{e}^{-1}\left(B\right) \supset\cup_{j=1}^{m}f_{e}^{-1}\left(A_{i_{j}}\right)$.

Finally, we reach $\bar{f}_{e}^{-1}\left(B\right)=\cup_{j=1}^{m}f_{e}^{-1}\left(A_{i_{j}}\right)$, which concludes our proof because $f_e$ is a measurable function and $A_{i_j}$ are measurable sets. 

\end{proof}

\begin{theorem}
\textbf{(Theorem~\ref{thm:trainable_single} in the main paper)}
\label{thm:trainable_single_apd}
If we seek $f_{d}$ and $f_{e}$ in a family
with infinite capacity (e.g., the family of all measurable functions),
the the two OPs of interest in (\ref{eq:single_level_apd}) is equivalent to the following OP
\begin{equation}
\min_{\mathbb{P}_{c,\pi}} \min_{f_{d},f_{e}}
\mathbb{E}_{x\sim\mathbb{P}_{x}}\left[\ell\left(f_{d}\left(Q_C\left(f_{e}\left(x\right)\right)\right),f_y(x)\right)\right]+\lambda\mathcal{W}_{d_{z}}\left(f_{e}\#\mathbb{P}_{x},\mathbb{P}_{c,\pi}\right),
\label{eq:reconstruct_form_continuous_apd}
\end{equation}

where and the parameter $\lambda>0$ and $Q_C(z)  = \text{argmin}{}_{c\in C}d_{z}\left(z,c\right)$
\end{theorem}

\begin{proof}

Given the optimal solution $C^{*1},\pi^{*1},f_{d}^{*1}$, and $f_{e}^{*1}$
of the OP in (\ref{eq:single_level_apd}), we conduct the
optimal solution for the OP in (\ref{eq:appendix_reconstruct_form.}). Let
us conduct $C^{*2}=C^{*1},f_{d}^{*2}=f_{d}^{*1}$. We next define
$\bar{f}_{e}^{*2}\left(x\right)=\text{argmin}_{c}d_{z}\left(f_{e}^{*1}\left(x\right),c\right)=Q_{C^{*1}}\left(f_{e}^{*1}\left(x\right)\right) = Q_{C^{*2}}\left(f_{e}^{*1}\left(x\right)\right)$. We prove that $C^{*2}, \pi^{*2}, f_d^{*2}$, and $\bar{f}_e^{*2}$ are optimal solution of the OP in (\ref{eq:appendix_reconstruct_form.}). By this definition, we yield $\bar{f}_{e}^{*2}\#\mathbb{P}_{x}=\mathbb{P}_{c^{*2},\pi^{*2}}$
and hence $\mathcal{W}_{d_{z}}\left(\bar{f}_{e}^{*2}\#\mathbb{P}_{x},\mathbb{P}_{c^{*2},\pi^{*2}}\right)=0$. Therefore, we need to verify the following:


(i) $\bar{f}_e^{*2}$ is a Borel-measurable function.

(ii) 
Given a feasible solution $C, \pi, f_d$, and $\bar{f}_e$ of (\ref{eq:appendix_reconstruct_form.}), we have 
\begin{align}
\mathbb{E}_{x\sim\mathbb{P}_{x}}\left[\ell\left(f_{d}^{*2}\left(\bar{f}_{e}^{*2}\left(x\right)\right),f_y(x)\right)\right] & \leq\mathbb{E}_{x\sim\mathbb{P}_{x}}\left[\ell\left(f_{d}\left(\bar{f}_{e}\left(x\right)\right),f_y(x)\right)\right]. \label{eq:ii}
\end{align}

 We derive as 
\begin{align}
&\mathbb{E}_{x\sim\mathbb{P}_{x}}\left[\ell\left(f_{d}^{*2}\left(\bar{f}_{e}^{*2}\left(x\right)\right),f_y(x)\right)\right] +\lambda\mathcal{W}_{d_{z}}\left(\bar{f}_{e}^{*2}\#\mathbb{P}_{x},\mathbb{P}_{c^{*2},\pi^{*2}}\right)
\nonumber\\
& = \mathbb{E}_{x\sim\mathbb{P}_{x}}\left[\ell\left(f_{d}^{*2}\left(\bar{f}_{e}^{*2}\left(x\right)\right),x\right)\right]
\nonumber\\
& = \mathbb{E}_{x\sim\mathbb{P}_{x}}\left[\ell\left(f_{d}^{*1}\left(Q_{C^{*2}}\left(f_{e}^{*1}\left(x\right)\right)\right),f_y(x)\right)\right]\nonumber\\
& = \mathbb{E}_{x\sim\mathbb{P}_{x}}\left[\ell\left(f_{d}^{*1}\left(Q_{C^{*1}}\left(f_{e}^{*1}\left(x\right)\right)\right),f_y(x)\right)\right]
\nonumber\\
& \leq \mathbb{E}_{x\sim\mathbb{P}_{x}}\left[\ell\left(f_{d}^{*1}\left(Q_{C^{*1}}\left(f_{e}^{*1}\left(x\right)\right)\right),f_y(x)\right)\right]+\lambda\mathcal{W}_{d_{z}}\left(f_{e}^{*1}\#\mathbb{P}_{x},\mathbb{P}_{c^{*1},\pi^{*1}}\right).
\label{eq:1}
\end{align}

Moreover,  because $\bar{f}_e\#\mathbb{P}_x = \mathbb{P}_{c,\pi}$ which is a discrete distribution over the set of codewords $C$, we obtain $Q_C(\bar{f}_e(x)) = \bar{f}_e(x)$. Note that $C, \pi, f_d$, and $\bar{f}_e$ is also a feasible solution of (\ref{eq:single_level_apd}) because $\bar{f}_e$ is also a specific encoder mapping from the data space to the latent space, we achieve  

\begin{align*}
& \mathbb{E}_{x\sim\mathbb{P}_{x}}\left[\ell\left(f_{d}\left(Q_{C}\left(\bar{f}_{e}\left(x\right)\right)\right),f_y(x)\right)\right]+\lambda\mathcal{W}_{d_{z}}\left(\bar{f}_{e}\#\mathbb{P}_{x},\mathbb{P}_{c,\pi}\right)\nonumber\\
& \geq\mathbb{E}_{x\sim\mathbb{P}_{x}}\left[\ell\left(f_{d}^{*1}\left(Q_{C^{*1}}\left(\bar{f}_{e}^{*1}\left(x\right)\right),f_y(x)\right)\right)\right]+\lambda\mathcal{W}_{d_{z}}\left(\bar{f}_{e}^{*1}\#\mathbb{P}_{x},\mathbb{P}_{c^{*1},\pi^{*1}}\right).
\end{align*}

Noting that $\bar{f}_e\#\mathbb{P}_x = \mathbb{P}_{c,\pi}$ and $Q_C(\bar{f}_e(x)) = \bar{f}_e(x)$, we arrive at
\begin{align}
& \mathbb{E}_{x\sim\mathbb{P}_{x}}\left[\ell\left(f_{d}\left(\bar{f}_{e}\left(x\right)\right),f_y(x)\right)\right]
\nonumber\\
& \geq\mathbb{E}_{x\sim\mathbb{P}_{x}}\left[\ell\left(f_{d}^{*1}\left(Q_{C^{*1}}\left(\bar{f}_{e}^{*1}\left(x\right)\right)\right),f_y(x)\right)\right]+\lambda\mathcal{W}_{d_{z}}\left(\bar{f}_{e}^{*1}\#\mathbb{P}_{x},\mathbb{P}_{c^{*1},\pi^{*1}}\right). \label{eq:2}
\end{align}
Combining the inequalities in (\ref{eq:1}) and (\ref{eq:2}), we obtain Inequality (\ref{eq:ii}) as

\begin{align}
\mathbb{E}_{x\sim\mathbb{P}_{x}}\left[\ell\left(f_{d}^{*2}\left(\bar{f}_{e}^{*2}\left(x\right)\right),f_y(x)\right)\right]
& \leq\mathbb{E}_{x\sim\mathbb{P}_{x}}\left[\ell\left(f_{d}\left(\bar{f}_{e}\left(x\right)\right),f_y(x)\right)\right]. 
\end{align}
This concludes our proof.

\end{proof}

\subsection{Proof Lemma~\ref{cor:distortion_data} in the main paper: Accuracy Trade-off}
\label{sec:proof-2}

We first demonstrate the equivalence between the optimization problem in Eq. (\ref{eq:single_level_apd}) and the Wasserstein distance between the prediction distribution $f_d\#\mathbb{P}_{c,\pi}$ and the label distribution $f_y\#\mathbb{P}_x$, as outlined in Proposition 1.3. Following this, we relate this result to a clustering perspective (Accuracy Trade-off Lemma).

\begin{proposition}
\label{thm:appendix_reconstruct_form} We can equivalently reformulate the OP in Eq. (\ref{eq:single_level_apd}) as minimizing the Wasserstein distance between the prediction distribution $f_d\#\mathbb{P}_{c,\pi}$ and the label distribution $f_y\#\mathbb{P}_x$:

\begin{equation}
\min_{C,\pi,f_{d}}\min_{\bar{f}_{e}:\bar{f}_{e}\#f_y\#\mathbb{P}_{x}=\mathbb{P}_{c,\pi}}\mathbb{E}_{x\sim\mathbb{P}_{x},c\sim\bar{f}_{e}\left(x\right)}\left[\ell\left(f_{d}\left(c\right),x\right)\right]=\min_{C,\pi,f_{d}}\mathcal{W}_{\ell}\left(f_{d}\#\mathbb{P}_{c,\pi},f_y\#\mathbb{P}_{x}\right),\label{eq:appendix_reconstruct_form.}
\end{equation}.
\end{proposition}

\begin{proof}

We first prove that 
\begin{equation}
\mathcal{W}_{\ell}\left(f_{d}\#\mathbb{P}_{c,\pi},f_y\#\mathbb{P}_{x}\right)=\min_{\bar{f}_{e}:\bar{f}_{e}\#f_y\#\mathbb{P}_{x}=\mathbb{P}_{c,\pi}}\mathbb{E}_{x\sim\mathbb{P}_{x},c\sim\bar{f}_{e}\left(x\right)}\left[\ell\left(f_{d}\left(c\right),f_y(x)\right)\right],\label{eq:appendix_ws_distance_latent}
\end{equation}
where $\bar{f}_{e}$ is a \textbf{stochastic discrete} encoder mapping
data example $x$ directly to the codebooks.

Let $\bar{f}_{e}$ be a \textbf{stochastic discrete} encoder such
that $\bar{f}_{e}\#\mathbb{P}_{x}=\mathbb{P}_{c,\pi}$ (i.e., $x\sim\mathbb{P}_{x}$
and $c\sim\bar{f}_{e}\left(x\right)$ implies $c\sim\mathbb{P}_{c,\pi}$).
We consider $\gamma_{d,c}$ as the joint distribution of $\left(x,c\right)$
with $x\sim\mathbb{P}_{x}\text{ and }c\sim\bar{f}_{e}\left(x\right)$.
We also consider $\gamma_{fc,d}$ as the joint distribution including
$\left(y,y'\right)\sim\gamma_{fc,d}$ where 
$x\sim\mathbb{P}_{x},$$c\sim\bar{f}_{e}\left(x\right)$, $y=f_y(x)$
and $y'=f_{d}\left(c\right)$. This follows that $\gamma_{fc,d}\in\Gamma\left(f_{d}\#\mathbb{P}_{c,\pi},f_y\#\mathbb{P}_{x}\right)$
which admits $f_{d}\#\mathbb{P}_{c,\pi}$ and $f_y\#\mathbb{P}_{x}$ as
its marginal distributions. We also have:
\begin{align*}
\mathbb{E}_{x\sim\mathbb{P}_{x},c\sim\bar{f}_{e}\left(x\right)}\left[\ell\left(f_{d}\left(c\right),f_y(x)\right)\right] & =\mathbb{E}_{(x,c)\sim\gamma_{d,c}}\left[\ell\left(f_{d}\left(c\right),f_y(x)\right)\right]\overset{(1)}{=}\mathbb{E}_{\left(y,y'\right)\sim\gamma_{fc,d}}\left[\ell\left(y,y'\right)\right]\\
 & \geq\min_{\gamma_{fc,d}\in\Gamma\left(f_{d}\#\mathbb{P}_{c,\pi},f_y\#\mathbb{P}_{x}\right)}\mathbb{E}_{\left(y,y'\right)\sim\gamma_{fc,d}}\left[\ell\left(y,y'\right)\right]\\
 & =\mathcal{W}_{\ell}\left(f_{d}\#\mathbb{P}_{c,\pi},f_y\#\mathbb{P}_{x}\right).
\end{align*}

Note that we have the equality in (1) due to $\left(f_y,f_{d}\right)\#\gamma_{d,c}=\gamma_{fc,d}$ (both $f_y$ and $f_{d}$ are deterministic functions). 

Therefore, we reach
\[
\min_{\bar{f}_{e}:\bar{f}_{e}\#\mathbb{P}_{x}=\mathbb{P}_{c,\pi}}\mathbb{E}_{x\sim\mathbb{P}_{x},c\sim\bar{f}_{e}\left(x\right)}\left[\ell\left(f_{d}\left(c\right),f_y(x)\right)\right]\geq\mathcal{W}_{\ell}\left(f_{d}\#\mathbb{P}_{c,\pi},f_y\#\mathbb{P}_{x}\right).
\]

Let $\gamma_{fc,d}\in\Gamma\left(f_y\#\mathbb{P}_{x}, f_{d}\#\mathbb{P}_{c,\pi}\right)$.
Let $\gamma_{fc,c}\in\Gamma\left(f_{d}\#\mathbb{P}_{c,\pi},\mathbb{P}_{c,\pi}\right)$
be a deterministic coupling such that $c\sim\mathbb{P}_{c,\pi}$ and
$y=f_{d}\left(c\right)$ imply $\left(c,y\right)\sim\gamma_{c,fc}$. Using
the gluing lemma (see Lemma 5.5 in \citep{santambrogio2015optimal}), there exists a joint distribution $\alpha\in\Gamma\left(\mathbb{P}_{c,\pi},f_{d}\#\mathbb{P}_{c,\pi},f_y\#\mathbb{P}_{x}\right)$
which admits $\gamma_{fc,d}$ and $\gamma_{fc,c}$ as the corresponding
joint distributions. By denoting $\gamma_{d,c}\in\Gamma\left(f_y\#\mathbb{P}_{x},\mathbb{P}_{c,\pi}\right)$
as the marginal distribution of $\alpha$ over $f_y\#\mathbb{P}_{x},\mathbb{P}_{c,\pi}$,
we then have
\begin{align*}
\mathbb{E}_{\left(y,y'\right)\sim\gamma_{fc,d}}\left[\ell\left(y,y'\right)\right] & =\mathbb{E}_{\left(c,y',y\right)\sim\alpha}\left[\ell\left(y,y'\right)\right]\\
&=\mathbb{E}_{\left(c,y\right)\sim\gamma_{d,c},y'=f_{d}\left(c\right)}\left[\ell\left(y,y'\right)\right]\\
 & =\mathbb{E}_{\left(c,y\right)\sim\gamma_{d,c}}\left[\ell\left(f_{d}\left(c\right),y\right)\right]\\
 &=\mathbb{E}_{x\sim\mathbb{P}_{x},c\sim\bar{f}_{e}\left(x\right)}\left[\ell\left(f_{d}\left(c\right),f_y(x)\right)\right].\\
 & \geq\min_{\bar{f}_{e}:\bar{f}_{e}\#\mathbb{P}_{x}=\mathbb{P}_{c,\pi}}\mathbb{E}_{x\sim\mathbb{P}_{x},c\sim\bar{f}_{e}\left(x\right)}\left[\ell\left(f_{d}\left(c\right),f_y(x)\right)\right],
\end{align*}
where $\bar{f}_e(x) = \gamma_{d,c}(\cdot \mid x)$.

This follows that
\begin{align*}
\mathcal{W}_{\ell}\left(f_{d}\#\mathbb{P}_{c,\pi},f_y\#\mathbb{P}_{x}\right) & =\min_{\gamma_{fc,d}\in\Gamma\left(f_{d}\#\mathbb{P}_{c,\pi},f_y\#\mathbb{P}_{x}\right)}\mathbb{E}_{\left(y,y'\right)\sim\gamma_{fc,d}}\left[\ell\left(y,y'\right)\right]\\
 & \geq\min_{\bar{f}_{e}:\bar{f}_{e}\#\mathbb{P}_{x}=\mathbb{P}_{c,\pi}}\mathbb{E}_{x\sim\mathbb{P}_{x},c\sim\bar{f}_{e}\left(x\right)}\left[\ell\left(f_{d}\left(c\right),f_y(x)\right)\right].
\end{align*}

This completes the proof for the equality in Eq. (\ref{eq:appendix_ws_distance_latent}),
which means that $\min_{C,\pi,f_{d}}\mathcal{W}_{\ell}\left(f_{d}\#\mathbb{P}_{c,\pi},f_y\#\mathbb{P}_{x}\right)$ is
equivalent to 
\begin{equation}
\min_{C,\pi,f_{d}}\min_{\bar{f}_{e}:\bar{f}_{e}\#\mathbb{P}_{x}=\mathbb{P}_{c,\pi}}\mathbb{E}_{x\sim\mathbb{P}_{x},c\sim\bar{f}_{e}\left(x\right)}\left[\ell\left(f_{d}\left(c\right),f_y(x)\right)\right],\label{eq:appendix_reconstruct_form.-1-1}
\end{equation}

We now further prove the above OP is equivalent to
\begin{equation}
\min_{C,\pi,f_{d}}\min_{\bar{f}_{e}:\bar{f}_{e}\#\mathbb{P}_{x}=\mathbb{P}_{c,\pi}}\mathbb{E}_{x\sim\mathbb{P}_{x}}\left[\ell\left(f_{d}\left(\bar{f}_{e}\left(x\right)\right),f_y(x)\right)\right],\label{eq:appendix_reconstruct_form.-2}
\end{equation}
where $\bar{f}_{e}$ is a \textbf{deterministic discrete} encoder
mapping data example $x$ directly to the codebooks.

It is obvious that the OP in (\ref{eq:appendix_reconstruct_form.-2}) is special
case of that in (\ref{eq:appendix_reconstruct_form.-1-1}) when we limit to
search for deterministic discrete encoders. Given the optimal solution
$C^{*1},\pi^{*1},f_{d}^{*1}$, and $\bar{f}_{e}^{*1}$ of the OP in
(\ref{eq:appendix_reconstruct_form.-1-1}), we show how to construct the optimal
solution for the OP in (\ref{eq:appendix_reconstruct_form.-2}). Let us construct
$C^{*2}=C^{*1}$, $f_{d}^{*2}=f_{d}^{*1}$. Given $x\sim\mathbb{P}_{x}$,
let us denote $\bar{f}_{e}^{*2}\left(x\right)=\text{argmin}{}_{c}\ell\left(f_{d}^{*2}\left(c\right),f_y(x)\right)$.
Thus, $\bar{f}_{e}^{*2}$ is a deterministic discrete encoder mapping
data example $x$ directly to the codebooks. We define $\pi_{k}^{*2}=Pr\left(\bar{f}_{e}^{*2}\left(x\right)=c_{k}:x\sim\mathbb{P}_{x}\right),k=1,...,K$,
meaning that $\bar{f}_{e}^{*2}\#\mathbb{P}_{x}=\mathbb{P}_{c^{*2},\pi^{*2}}$.
From the construction of $\bar{f}_{e}^{*2}$, we have 
\[
\mathbb{E}_{x\sim\mathbb{P}_{x}}\left[\ell\left(f_{d}^{*2}\left(\bar{f}_{e}^{*2}\left(x\right)\right),f_y(x)\right)\right]\leq\mathbb{E}_{x\sim\mathbb{P}_{x},c\sim\bar{f}_{e}^{*1}\left(x\right)}\left[\ell\left(f_{d}^{*1}\left(c\right),f_y(x)\right)\right].
\]

Furthermore, because $C^{*2},\pi^{*2},f_{d}^{*2},\text{and}\bar{f}_{e}^{*2}$
are also a feasible solution of the OP in (\ref{eq:appendix_reconstruct_form.-2}),
we have
\[
\mathbb{E}_{x\sim\mathbb{P}_{x}}\left[\ell\left(f_{d}^{*2}\left(\bar{f}_{e}^{*2}\left(x\right)\right),f_y(x)\right)\right]\geq\mathbb{E}_{x\sim\mathbb{P}_{x},c\sim\bar{f}_{e}^{*1}\left(x\right)}\left[\ell\left(f_{d}^{*1}\left(c\right),f_y(x)\right)\right].
\]

This means that 
\[
\mathbb{E}_{x\sim\mathbb{P}_{x}}\left[\ell\left(f_{d}^{*2}\left(\bar{f}_{e}^{*2}\left(x\right)\right),f_y(x)\right)\right]=\mathbb{E}_{x\sim\mathbb{P}_{x},c\sim\bar{f}_{e}^{*1}\left(x\right)}\left[\ell\left(f_{d}^{*1}\left(c\right),f_y(x)\right)\right],
\]
and $C^{*2},\pi^{*2},f_{d}^{*2},\text{and}\bar{f}_{e}^{*2}$ are also
the optimal solution of the OP in (\ref{eq:appendix_reconstruct_form.-2}).

\end{proof}



\begin{lemma} \textbf{(Lemma~\ref{cor:distortion_data} in the main paper)}
\label{cor:distortion_data_apd} Let $C^{*}=\left\{ c_{m}^{*}\right\} _{k=1}^M,\pi^{*}$,
and $f_{d}^{*}$ be the optimal solution of the OP in Eq. (\ref{eq:single_level_apd}), then $C^{*}=\left\{ c_{m}^{*}\right\} _{k},\pi^{*}$,
and $f_{d}^{*}$ are also the optimal solution of the following OP:

\begin{enumerate}
    \item $
\epsilon^{*}_{M} = \min_{f_{d}}\min_{\pi}\min_{\sigma\in\Sigma_{\pi}}\mathbb{E}_{x\sim \mathbb{P}_x}\left [\ell\left(f_y(x),f_{d}\left(c_{\sigma(x)}\right)\right) \right ]$
\item $\epsilon^{*}_M \geq \epsilon^{*}_{M+1} \text{ } \forall M$
\item $\epsilon^{*}_M \rightarrow 0 \text{ as } M\rightarrow\infty$ 
\end{enumerate}

where $\Sigma_{\pi}$ is the set of assignment functions $\sigma:\mathcal{X} \rightarrow\left\{ 1,...,M\right\} $
such that $ \mathbb{P}_x\left(\sigma^{-1}\left(m\right)\right),m=1,...,M$
are proportional to $\pi_{k},m=1,...,M$  (e.g., $\sigma$ is the nearest assignment: $\sigma^{-1}\left(m\right) = \{x\mid \bar{f_{e}}\left(x\right) = c_m, m=\text{argmin} _{m}d_{z}\left(f_{e}\left(x\right),c_{m}\right)\}$ is set of latent representations which are quantized to $m^{th}$ codeword).
\end{lemma}

\begin{proof}(1)

Based on the result of Proposition~\ref{thm:appendix_reconstruct_form}, the OP in Eq.~(\ref{eq:single_level_apd}) can be rewritten as:

\begin{equation*}
\min_{C,\pi}\min_{f^{d}}\min_{\bar{f}_{e}:\bar{f}_{e}\#f_y\#\mathbb{P}_{x}=\mathbb{P}_{c,\pi}}\mathbb{E}_{x\sim\mathbb{P}_{x},c\sim\bar{f}_{e}\left(x\right)}\left[\ell\left(f_{d}\left(c\right),x\right)\right]=\min_{C,\pi}\min_{f^{d}}\mathcal{W}_{\ell}\left(f_{d}\#\mathbb{P}_{c,\pi},f_y\#\mathbb{P}_{x}\right)
\end{equation*}

By definition, it is clear that
\[
f_{d}\#\mathbb{P}_{c,\pi}=\sum_{k=1}^{K}\pi_{k}\delta_{f_{d}\left(c_{k}\right)}.
\]

Therefore, the  OP in right hand side can be rewritten as:
\begin{equation}
\min_{C,\pi}\min_{f^{d}}\mathcal{W}_{\ell}\left(f_y\#\mathbb{P}_x,\sum_{k=1}^{K}\pi_{k}\delta_{f_{d}\left(c_{k}\right)}\right).\label{eq:appendix_push_forward-1}
\end{equation}

By using the Monge definition, we have 
\begin{align*}
\mathcal{W}_{\ell}\left(f_y\#\mathbb{P}_x,\sum_{k=1}^{K}\pi_{k}\delta_{f_{d}\left(c_{k}\right)}\right) & =\min_{T:T\#f_y\#\mathbb{P}_{x}=f_{d}\#\mathbb{P}_{c,\pi}}\mathbb{E}_{x\sim\mathbb{P}_{x}}\left[\ell\left(f_y(x),T\left(f_y\left(x\right)\right)\right)\right]
\end{align*}

Since $T\#f_y\#\mathbb{P}_{x}=f_{d}\#\mathbb{P}_{c,\pi}$, $T\left(x\right)=f_{d}\left(c_{k}\right)$
for some $k$. Additionally, $\left|T^{-1}\left(f_d(c_{k})\right)\right|,k=1,...,K$
are proportional to $\pi_{k},k=1,...,K$. Denote $\sigma:\mathcal{X} \rightarrow\left\{ 1,...,K\right\} $
such that $T\left(f_y(x)\right)=f_d(c_{\sigma\left(x\right)}),\forall x\in \mathcal{X}$,
we have $\sigma\in\Sigma_{\pi}$. It follows that 
\[
\mathcal{W}_{\ell}\left(f_y\#\mathbb{P}_x,\sum_{k=1}^{K}\pi_{k}\delta_{f_{d}\left(c_{k}\right)}\right)=\min_{\sigma\in\Sigma_{\pi}}\mathbb{E}_{x\sim \mathbb{P}_x}\left [\ell\left(f_y(x),f_d\left(c_{\sigma\left(x\right)}\right)\right)\right ].
\]

Finally, the optimal solution of the OP in Eq. (\ref{eq:single_level_apd})
is equivalent to
\[
\min_{f_{d}}\min_{\pi}\min_{\sigma\in\Sigma_{\pi}}\mathbb{E}_{x\sim \mathbb{P}_x}\left [\ell\left(f_y(x),f_{d}\left(c_{\sigma(n)}\right)\right) \right ],
\]
which directly implies the conclusion.
\end{proof}

\begin{proof}(2)-(3)
Given $f^*_e$, $C_M^{*}=\left\{ c_{m}^{*}\right\}_{m=1}^M ,\pi^{*}$ and $f_{d}^{*}$ be the optimal solution of the OP in Eq. (\ref{eq:single_level_apd}) w.r.t to empirical training data $\mathbb{P}^n_{x}$, then the optimal loss is:
\begin{align}
    \epsilon^*_M &=\mathbb{E}_{x\sim\mathbb{P}_{x}}\left[\ell\left(f^*_{d}\left(Q_C\left (f^*_{e}\left(x\right) \right )\right),f_y(x)\right)\right] \nonumber\\
    &=\int_{\mathcal{X}}\mathbb{P}(x)\ell\left(f^*_{d}\left(Q_C\left (f^*_{e}\left(x\right) \right )\right),f_y(x)\right) dx
\end{align}  

We construct a solution with a codebook size of \( (M+1) \) as follows.

\begin{itemize}
    \item First choose $x_{\textbf{max}}=\arg\min_{x\sim\mathbb{P}_x}\ell\left(f_{d}\left(\bar{f}_{e}\left(x\right)\right),f_y(x)\right)$
    \item We construct $C_{M+1}=C_M^{*}\cup \{c_{M+1}\}=\left\{ c_{m}^{*}\right\}_{m=1}^M$ $\{c_{M+1}\}$ such that $f_{d}^{*}\left ( c \right )=f_y(x_{\textbf{max}})$
    \item Then we construct $f_e$ such that $Q_c(f_e(x_{\textbf{max}}))=c_{M+1}$ and $f_e(x)=f^*_e(x)$ for all $x\neq x_{\textbf{max}}$. 
\end{itemize}

Then given $f_e$, $C_{M+1}$, $f^*_d$, then loss w.r.t to $\mathbb{P}_x$ is:

\begin{align*}
\epsilon_{M+1}&=\mathbb{E}_{x\sim\mathbb{P}_{x}}\left[\ell\left(f^*_d\left (Q_C\left ( f_e\left ( x\right ) \right )\right ),f_y(x)\right)\right]\\
 &=\int_{\mathcal{}}\mathbb{P}(x)\ell\left(f^*_{d}\left(Q_C\left (f_{e}\left(x\right) \right )\right),f_y(x)\right) dx\\
     &=\int_{\mathbb{X} \setminus\{x_{\textbf{max}}\} }\mathbb{P}(x)\ell\left(f^*_{d}\left(Q_C\left (f_{e}\left(x\right) \right )\right),f_y(x)\right) dx + \mathbb{P}(x_{\textbf{max}})\ell\left(f^*_{d}\left(Q_C\left (f_{e}\left(x_{\textbf{max}}\right) \right )\right),f_y(x)\right)\\
     &=\int_{\mathbb{X} \setminus\{x_{\textbf{max}}\} }\mathbb{P}(x)\ell\left(f^*_{d}\left(Q_C\left (f^*_{e}\left(x\right) \right )\right),f_y(x)\right) dx + \mathbb{P}(x_{\textbf{max}})\ell\left(f^*_{d}\left(Q_C\left (f_{e}\left(x_{\textbf{max}}\right) \right )\right),f_y(x)\right)\\
     &=\int_{\mathbb{X} \setminus\{x_{\textbf{max}}\} }\mathbb{P}(x)\ell\left(f^*_{d}\left(Q_C\left (f^*_{e}\left(x\right) \right )\right),f_y(x)\right) dx \\
     & \leq \int_{\mathbb{X} \setminus\{x_{\textbf{max}}\} }\mathbb{P}(x)\ell\left(f^*_{d}\left(Q_C\left (f^*_{e}\left(x\right) \right )\right),f_y(x)\right) dx + \mathbb{P}(x_{\textbf{max}})\ell\left(f^*_{d}\left(Q_C\left (f^*_{e}\left(x_{\textbf{max}}\right) \right )\right),f_y(x)\right)\\ 
     &=\epsilon^*_{M}
\end{align*}

Since \( f_e \), \( C_{M+1} \), and \( f^*_d \) are feasible solutions of the OP in Eq. (\ref{eq:single_level_apd}) with a codebook size of \( (M+1) \), we have \( \epsilon_{M+1} \geq \epsilon^*_{M+1} \). Hence, we reach \( \epsilon^*_{M+1} \leq \epsilon^*_{M} \). 

Additionally, the equation holds when \( \mathbb{P}(x_{\textbf{max}})\ell\left(f^*_{d}\left(Q_C\left (f^*_{e}\left(x_{\textbf{max}}\right) \right)\right), f_y(x)\right) = 0 \), which implies \( \epsilon^*_{M} = 0 \), directly implying the conclusion (2) and (3).
\end{proof}

\subsection{Proof Corollary~\ref{cor:sample_complexity} in the main paper: Sample Complexity}
\label{sec:proof-3}
We outline the setup related to Corollary 3.3 before proceeding with the proof. 


Let the training set be $S=\left\{ x_{1},...,x_{N}\right\}$ consisting of $N$ sample $x_i\sim \mathbb{P}_x$ and define the empirical measure $\mathbb{P}^N_{x}$ by placing mass $N^{-1}$ at each of members of $S$ i.e., $\mathbb{P}^N_{x}= \sum_{i=1}^{N}\frac{1}{N}\delta_{x_{i}}$.
Given $C^{*}=\left\{ c_{m}^{*}\right\}_{m=1}^M ,\pi^{*}$, $f_e^*$, and $f_{d}^{*}$ be the optimal solution of the OP in Eq. (\ref{eq:single_level_apd}) w.r.t to empirical training data $\mathbb{P}^N_{x}$, the empirical loss w.r.t to distribution $\mathbb{P}^N_x$ is defined as:
\begin{equation}
\mathcal{L}\left(\mathbb{P}^N_x\right) =\mathbb{E}_{x\sim \mathbb{P}^N_x}\left [ \ell\left(f_y(x),f_d^*(Q_C(f_e^{*}(x)))\right)\right ],\label{eq:true_loss}
\end{equation}
Notably, Lemma \ref{cor:distortion_data} establishes that solving the optimization problem (OP) in Eq. (\ref{eq:single_level}) is equivalent to solving a clustering problem in the output space. However, discrete representations exhibit distinct characteristics. Specifically, defining $\bar{f}_y(x)=f_d^*(\underset{c''\in C}{\text{argmin}}\ell_y(x, f^*_d(c'')))$, we obtain:
\begin{align}
\mathcal{L}\left(\mathbb{P}^N_x\right) =&\mathbb{E}_{x\sim \mathbb{P}^N_x}\left [ \ell\left(f_y(x),f_d^*(\bar{f}_e(x))\right)\right ]\nonumber\\
=& \underset{\text{Clustering Loss: }\mathcal{L}_C\left(\mathbb{P}^N_x\right)}{\underbrace{\mathbb{E}_{x\sim \mathbb{P}^N_x}\left [\ell\left(\bar{f}_y(x),f_y(x)\right)\right ]}}+ \underset{\text{Assignment Discrepancy Loss: }\mathcal{L}_A\left(\mathbb{P}^N_x\right)}{\underbrace{\mathbb{E}_{x\sim \mathbb{P}^N_x}\left [\ell\left(f_{d}^*\left( (\bar{f}_{e}^*\left(x\right) \right),f_y(x)\right) -\ell\left(\bar{f}_y(x),f_y(x)\right)\right ]}}\nonumber\\
=&\mathcal{L}_C\left(\mathbb{P}^N_x\right)+\mathcal{L}_A\left(\mathbb{P}^N_x\right) \nonumber
\end{align}
The overall error is decomposed into two components:

\textit{Clustering Loss}: The fist term $\ell\left(\bar{f}_y(x),f_y(x)\right)$ represents the distance between the label $f_y(x)$ and its nearest centroid in the output space, which is determined by mapping the codebook to the output space.

\textit{Assignment Discrepancy Loss}: It can be seen that $\bar f_e^*(x)$ is the nearest assignment on latent space while $\bar{f}_y(x)=f_d^*(\underset{c''\in C}{\text{argmin}}\ell_y(x, f^*_d(c'')))$ is the nearest assignment on output space. The second term i.e., the discrepancy between $\ell\left(f_{d}^*\left( (\bar{f}_{e}^*\left(x\right) \right),f_y(x)\right) -\ell\left(\bar{f}_y(x),f_y(x)\right)$ arises from the mismatch between assignments in the latent and output spaces. Specifically, in Figure \ref{fig:assignment}, for sample 
, the decoded centroid $f_d(c_3)$ in the latent space may not be the closest centroid in the output space.


Following that, the general loss is defined similarly:
\begin{align}
\mathcal{L}\left(\mathbb{P}_x\right) =\mathcal{L}_C\left(\mathbb{P}_x\right)+\mathcal{L}_A\left(\mathbb{P}_x\right) \nonumber
\end{align}
The discrepancy between the empirical loss and the general loss can be bounded as follows:
\begin{align}
\left |\mathcal{L}\left(\mathbb{P}_x^N\right) - \mathcal{L}\left(\mathbb{P}_x\right) \right |\nonumber&= \left | \mathcal{L}_C\left(\mathbb{P}^N_x\right)+\mathcal{L}_A\left(\mathbb{P}^N_x\right) - \mathcal{L}_C\left(\mathbb{P}_x\right)-\mathcal{L}_A\left(\mathbb{P}_x\right) \right |\nonumber\\
&\leq \left | \mathcal{L}_C\left(\mathbb{P}^N_x\right)-\mathcal{L}_C\left(\mathbb{P}_x\right)\right | +\left |\mathcal{L}_A\left(\mathbb{P}^N_x\right)-\mathcal{L}_A\left(\mathbb{P}_x\right) \right |\nonumber
\end{align}
The first term $\left | \mathcal{L}_C\left(\mathbb{P}^N_x\right)-\mathcal{L}_C\left(\mathbb{P}_x\right)\right |$ is the generalization bounds for k-mean problem, which is studied in ~\citep{telgarsky2013moment, bachem2017uniform}. In this study, our focus is on the second term, $ \left |\mathcal{L}_A\left(\mathbb{P}^N_x\right)-\mathcal{L}_A\left(\mathbb{P}_x\right) \right |$ , which examines the impact of discrete representations on the optimization problem.

We begin the proof by presenting Proposition~\ref{prop:discrete_concentration}, which is used to prove Corollary~\ref{cor:sample_complexity} in the main paper. 

\begin{proposition}[Concentration for Discrete Distributions \citep{hsu2012spectral, agarwal2019reinforcement}] \label{prop:discrete_concentration} Let $z$ be a discrete random variable that takes values in $\{1, \dots, d\}$, distributed according to $q$. We write $q$ as a vector where $q=\left [\mathbb P (z=j)\right ]_{j=1}^d$. Assume we have $n$ iid samples, and that our empirical estimate of $q$ is $\hat q = \left [\frac{1}{N}\sum_{i=1}^N\mathbf{1} [z_j=j]\right ]_{j=1}^d$. We have that $\forall \epsilon > 0$

\[ Pr \left( \|\hat q - q\|_2 \geq 1/\sqrt{N} + \epsilon \right) \leq e^{-N\epsilon^2}\]
which implies that 
\[ Pr \left( \|\hat q - q\|_1 \geq \sqrt{d}(1/\sqrt{N} + \epsilon) \right) \leq e^{-N\epsilon^2}\]
    
\end{proposition}


\begin{corollary} \textbf{Corollary~\ref{cor:sample_complexity} in the main paper}
        Given a dataset of size $N$, a loss function $\ell_y$ that is a proper metric and upper-bounded by a positive constant $L$, and an encoder class $\mathcal{F}_e$ (i.e., $\bar{f}_e\in\mathcal{F}_e$), then it holds with probability at least $1-\delta$ that
    \begin{equation}
        |\mathcal{L}_A\left(\mathbb{P}_x^N\right) - \mathcal{L}_A\left(\mathbb{P}_x\right)| \leq \epsilon
    \end{equation}
where $\epsilon = L{M}{\frac{\sqrt{\log 1/\delta + \log |\mathcal{F}_e|} + 1}{\sqrt{N}}} $.
\end{corollary}

\begin{proof}

\begin{align*}
|\mathcal{L}_A\left(\mathbb{P}_x^N\right) - \mathcal{L}_A\left(\mathbb{P}_x\right)|&=\left |\int_{\mathcal{X}}\left (\mathbb{P}^N(x) -\mathbb{P}(x)\right ) \left( \ell\left(f_{d}\left( (\bar{f}_{e}^*\left(x\right) \right),f_y(x)\right) -\ell\left(\bar{f}_y(x),f_y(x)\right)\right ) dx \right |\\
&\overset{(1)}{\leq}\left |\int_{\mathcal{X}}\left (\mathbb{P}^N(x) -\mathbb{P}(x)\right )  \ell\left(f_{d}\left( (\bar{f}_{e}^*\left(x\right) \right),\bar{f}_y(x)\right) dx \right |\\
&=\left |\int_{\mathcal{X}}(\mathbb{P}^N(x)  \ell\left(f_{d}\left( (\bar{f}_{e}^*\left(x\right) \right),\bar{f}_y(x)\right) dx -\int_{\mathcal{X}}\mathbb{P}(x)\ell\left(f_{d}\left( (\bar{f}_{e}^*\left(x\right) \right),\bar{f}_y(x)\right) dx\right |\\
&=|\mathcal{L}_A\left(\bar{f}_y, \mathbb{P}_x^N\right) - \mathcal{L}_A\left(\bar{f}_y,\mathbb{P}_x\right)|
\end{align*}

We have $\overset{(1)}{\leq}$ because:
\begin{itemize}
    \item Since  $\bar{f}_y(x)=f_d^*(\underset{c''\in C}{\text{argmin}}\ell_y(x, f^*_d(c'')))$, then $\ell\left(f_{d}^*\left( (\bar{f}_{e}^*\left(x\right) \right),f_y(x)\right) - \ell\left(\bar{f}_y(x),f_y(x)\right)\geq 0$ for all $x$.
    \item by triangle inequality $\ell\left(f_{d}^*\left( (\bar{f}_{e}^*\left(x\right) \right),\bar{f}_y(x)\right)\geq \ell\left(f_{d}^*\left( (\bar{f}_{e}^*\left(x\right) \right),f_y(x)\right) - \ell\left(\bar{f}_y(x),f_y(x)\right)$ .
    \item  Since $C^{*}=\left\{ c_{m}^{*}\right\}_{m=1}^M ,\pi^{*}$, $f_e^*$, and $f_{d}^{*}$ be the optimal solution of the OP in Eq. (\ref{eq:single_level_apd}) w.r.t to empirical training data $\mathbb{P}^N_{x}$, $\ell\left(f_{d}^*\left( (\bar{f}_{e}^*\left(x\right) \right),f_y(x)\right) - \ell\left(\bar{f}_y(x),f_y(x)\right)= 0$ for all $x$ such that $\mathbb{P}^N(x)>0$.
    \item $\left (\mathbb{P}(x)^N -\mathbb{P}(x)\right )<=0$ for all $x$ such that $\mathbb{P}^N(x)=0$.
\end{itemize}

Therefore,


\begin{align}
\mathcal{L}_A\left(\bar{f}_y,\mathbb{P}_x\right) &=\mathbb{E}_{x\sim \mathbb{P}_x}\left [ \ell\left(\bar{f}_y(x),f_d^*(\bar{f}_e(x))\right)\right ]\nonumber\\
& =\int_{\mathcal{X}}\mathbb{P}(x)\ell\left(f_{d}\left( (\bar{f}_{e}^*\left(x\right) \right),\bar{f}_y(x)\right) dx \nonumber\\
&=\int_{\mathcal{X}}\mathbb{P}(x)\sum_{c\in C}\mathbf{1}_{[\bar f_e^*(x)=c]} \ell(f_d^*(c), \bar{f}_y(x)) dx\nonumber\\
&=\sum_{c\in C}\int_{\mathcal{X}}\mathbb{P}(x)\mathbf{1}_{[\bar f_e^*(x)=c]} \ell(f_d^*(c), f_d^*(\min_{c''\in C}\ell_y(x, f^*_d(c'')))) dx\nonumber\\
&=\sum_{c\in C}\sum_{c'\in C}\int_{\mathcal{X}}\mathbb{P}(x)\mathbf{1}_{[\bar f_e^*(x)=c, \text{ }\underset{c''\in C}{\text{argmin}}\ell_y(x, f^*_d(c''))=c']} \ell(f_d^*(c), f_d^*(c')) dx\nonumber\\
\end{align}


Similarly, let $\mathbb{P}^N(x)=0$ $\forall x\notin S$, we have:
\begin{align}
\mathcal{L}_A\left(\bar{f}_y,\mathbb{P}^N_x\right) &=\sum_{c\in C}\sum_{c'\in C}\int_{\mathcal{X}}\mathbb{P}^N(x)\mathbf{1}_{[\bar f_e^*(x)=c, \text{ }\underset{c''\in C}{\text{argmin}}\ell_y(x, f^*_d(c''))=c']} \ell(f_d^*(c), f_d^*(c')) dx\nonumber
\end{align}

The discrepancy between the empirical loss and the general loss, arising from the mismatched assignment between the output space and the latent space, can be bounded as follows:

\begin{align}
&\left |\mathcal{L}\left(\bar{f}_y,\mathbb{P}_x^N\right) - \mathcal{L}\left(\bar{f}_y,\mathbb{P}_x\right) \right |\\
&=\left | \sum_{c\in C}\sum_{c'\in C}\int_{\mathcal{X}} \left ( \mathbb{P}^N(x)\mathbf{1}_{[\bar f_e^*(x)=c, \text{ }\underset{c''\in C}{\text{argmin}}\ell_y(x, f^*_d(c''))=c']} -\mathbb{P}(x)\mathbf{1}_{[\bar f_e^*(x)=c, \text{ }\underset{c''\in C}{\text{argmin}}\ell_y(x, f^*_d(c''))=c']}\right ) \ell(f_d^*(c), f_d^*(c'))  dx \right | \nonumber\\
&=\left | \sum_{c\in C}\sum_{c'\in C}\ell(f_d^*(c), f_d^*(c'))\int_{\mathcal{X}} \left ( \mathbb{P}^N(x)\mathbf{1}_{[\bar f_e^*(x)=c, \text{ }\underset{c''\in C}{\text{argmin}}\ell_y(x, f^*_d(c''))=c']} -\mathbb{P}(x)\mathbf{1}_{[\bar f_e^*(x)=c, \text{ }\underset{c''\in C}{\text{argmin}}\ell_y(x, f^*_d(c''))=c']}\right )  dx \right | \nonumber\\
&=\sum_{c\in C}\sum_{c'\in C}\left |  \ell(f_d^*(c), f_d^*(c'))   \right |\left |\int_{\mathcal{X}}  \mathbb{P}^N(x)\mathbf{1}_{[\bar f_e^*(x)=c, \text{ }\underset{c''\in C}{\text{argmin}}\ell_y(x, f^*_d(c''))=c']} -\mathbb{P}(x)\mathbf{1}_{[\bar f_e^*(x)=c, \text{ }\underset{c''\in C}{\text{argmin}}\ell_y(x, f^*_d(c''))=c']}\right | dx\nonumber\\
&=L\sum_{c\in C}\sum_{c'\in C}\left |\int_{\mathcal{X}}  \mathbb{P}^N(x)\mathbf{1}_{[\bar f_e^*(x)=c, \text{ }\underset{c''\in C}{\text{argmin}}\ell_y(x, f^*_d(c''))=c']} -\mathbb{P}(x)\mathbf{1}_{[\bar f_e^*(x)=c, \text{ }\underset{c''\in C}{\text{argmin}}\ell_y(x, f^*_d(c''))=c']}\right | dx\nonumber\\
&= L \|\hat P_{C} - P_{C}\|_1
\end{align}

where $\hat P_C$ and $P_C$ are the estimated and the true probability distribution vector of the product codebook $C\times C'\in \mathcal{C}\times\mathcal{C}$, i.e. $\hat{\mathbb{P}}_C=[\frac{1}{N}\sum_{i=1}^N\mathbf{1}_{[\bar f_e^*(x)=c, \text{ }\underset{c''\in C}{\text{argmin}}\ell_y(x, f^*_d(c''))=c']}]_{c,c'\in C}$ and $\mathbb{P}_C=[\mathbb{P} (\bar f_e^*(x)=c, \text{ }\underset{c''\in C}{\text{argmin}}\ell_y(x, f^*_d(c''))=c')]_{c,c'\in C}$.


    
Applying the concentration inequality from Proposition \ref{prop:discrete_concentration}, we have:

    \[ Pr \left( \|\hat{\mathbb{P}}_C - \mathbb{P}_C\|_1 \geq {M}(1/\sqrt{N} + \epsilon) \right) \leq e^{-N\epsilon^2},\]

Next, we apply the union bound to the above result across all optimal functions in the encoder class $\mathcal{F}_e$, which gives:

    \[ Pr \left(\exists \bar{f}_e \in  \mathcal{F}_e,  \|\hat{\mathbb{P}}_C - \mathbb{P}_C\|_1 \geq {M}(1/\sqrt{N} + \epsilon) \right) \leq |\mathcal{F}_e|e^{-N\epsilon^2},\]
    Let $\delta = |\mathcal{F}_e|e^{-N\epsilon^2}$, then with probability of at least $1-\delta$
    \begin{equation}
        Pr \left(  \|\hat{\mathbb{P}}_C - \mathbb{P}_C\|_1 \leq {M}(1/\sqrt{N} + \epsilon) \right) \geq 1-\delta, \quad \forall \bar{f}_e \in \mathcal{F}_e \label{eq:finite_samples}
    \end{equation}

Finally, combining this with (\ref{eq:finite_samples}), we have with probability at least \( 1 - \delta \):

    \[ |\mathcal{L}_A\left(\bar{f}_y,\mathbb{P}_x^N\right) - \mathcal{L}_A\left(\bar{f}_y,\mathbb{P}_x\right)|  \leq L{M}{\frac{\sqrt{\log 1/\delta + \log |\mathcal{F}_e|} + 1}{\sqrt{N}}} \]
\end{proof}



\subsection{Equivalence between $K$-class classification and Discrete Representation Framework}
\label{sec:proof-4}
\label{sec:kclass}
We consider the classifier on top of the encoder as a linear layer represented by the weight matrix $f_d = W = [w_k]_{k=1}^K$, where $w_k$ is the vector corresponding to class ``k". The prediction via softmax activation is given by:
$$
P(y \mid x) = \frac{\exp\left(f_e(x)^\top W_k\right)}{\sum_{j=1}^{K} \exp\left(f_e(x)^\top W_j\right)}
$$

Given that the loss of interest is the \textit{cross-entropy loss}, which is commonly used for classification tasks, the functions \( f_e \) and \( f_d \) are learned through the following optimization problem (OP):
\begin{align}
&\min_{W,f_{e}}   \underset{x\sim\mathbb{P}_{x}}{\mathbb{E}}\left[\ell\left(f_{d}\left(f_{e}\left(x\right)\right),f_y(x)\right)\right]\nonumber \\
&=\min_{W, f_{e}}   \underset{x\sim\mathbb{P}_{x}}{\mathbb{E}}\left[ f_y(x) \log \left (\frac{\exp\left(f_e(x)^\top W_{f_y(x)}\right)}{\sum_{j=1}^{K} \exp\left(f_e(x)^\top W_j\right)}\right )\right]
\label{eq:cls}
\end{align}
This loss reaches its optimal solution as \( f_e(x) = \lambda W_{f_y(x)} \) with \( \lambda \) as a coefficient, and \( f_e(x)^\top W_j=0 \) for all \( j \neq f_y(x) \). Since the optimal loss remains unchanged for any value of \( \lambda \), we can, without loss of generality, set \( \lambda = 1 \). 

The above optimal solution for \( f_e \) may be too restrictive. In fact, given \( W \), we only need to find \( f_e \) such that \( W_{f_y(x)} = Q_W(f_e(x))=\arg\max_{w \in W} d_z\left(f_e(x), w\right) \).

Given that \( f_e^* \) and \( W^* \) are the optimal solutions of the optimization problem in Eq.~(\ref{eq:cls}), then \( \bar{f}_e^*\#\mathbb{P}_x = \mathbb{P}_{c, \pi} \), where $\bar{f}_e^*=Q_W\circ f_e^*$ \( C = W*=\{w_k\}_{k=1}^K \), \( \pi = f_y\#\mathbb{P}_x \), \( \mathbb{P}_{c,\pi} = \sum_{k=1}^{K} \pi_k \delta_{c_k} \) and $f_d^*=W^*$ also the optimal solution of OP in Eq.~(\ref{eq:single_level_apd}):
\begin{align}
\min_{\mathbb{P}_{c,\pi},f_{d}}\min_{\bar{f}_{e}:\bar{f}_{e}\#\mathbb{P}_{x}=\mathbb{P}_{c,\pi}}   \underset{x\sim\mathbb{P}_{x}}{\mathbb{E}}\left[\ell\left(f_{d}\left(\bar{f}_{e}\left(x\right)\right),f_y(x)\right)\right] \nonumber
\end{align}

\section{Experimental Settings}
\label{apd:expsetting}

\subsection{Domain Generalization}
We adopt the training and evaluation protocol as described in the DomainBed benchmark \citep{gulrajani2020search}, including dataset splits, model selection on the validation set, and optimizer settings. To improve computational efficiency, as recommended by \citep{cha2021swad}, we restrict our HP search space. Particularly, we utilize the Adam optimizer with a learning rate of $5 \times 10^{-5}$, as detailed in \citep{gulrajani2020search}, without any adjustments for dropout or weight decay. The batch size is set to 32.

Our method's unique hyperparameter, the regularization term \( \lambda \), is optimized within the range of $[0.01, 0.1, 1.0]$, and the number of discrete representations \( \left | \mathcal{M} \right | \) is fixed at 16 times the number of classes.

SWAD-specific hyperparameters remain at their default settings, and the evaluation frequency is set to 300 for all datasets.

\subsection{Datasets}
To evaluate the effectiveness of the proposed method, we utilize five
datasets: PACS~\citep{li2017deeper}, VLCS~\citep{torralba2011unbiased},
 Office-Home~\citep{venkateswara2017deep}, Terra Incognita~\citep{beery2018recognition} and DomainNet~\citep{peng2019moment} which are the common DG benchmarks with multi-source domains.
\begin{itemize}
    \item \textbf{PACS}~\citep{li2017deeper}: 9991 images of seven classes in total, over four domains:Art\_painting (A), Cartoon (C), Sketches (S), and Photo (P). 
    
    \item \textbf{VLCS}~\citep{torralba2011unbiased}: five classes over four domains with a total of 10729 samples. The domains are defined by four image origins, i.e., images were taken from the PASCAL VOC 2007 (V), LabelMe (L), Caltech (C) and Sun (S) datasets.

    \item \textbf{Office-Home}~\citep{venkateswara2017deep}: 65 categories of 15,500 daily objects from 4 domains: Art, Clipart, Product (vendor website with white-background) and Real-World (real-object collected from regular cameras).
   
\end{itemize}
\subsection{State abstraction Reinforcement Learning}


We consider the Visual Gridworld environment from \citep{allen2021learning}. The underlying grid states is a $10\times 10$ grid structure, where the agent occupies individual squares and can navigate in four possible adjacent squares.  The agent receives the image observation from the environment of size $30\times30$ pixels, with its current location highlighted. The images are interfered with truncated Gaussian noise, making it crucial for an effective encoder to distinguish between the noise and the agent's true position for downstream tasks. See Figure (\ref{fig:gridworld}) for an illustration.

The learned encoder serves as a pre-trained feature extractor from image observation for training a DQN agent for downstream tasks. A random goal position is designated through out this training phase, and the agent receives a reward of (-1) for each step taken until it reaches the goal, upon which the episode stops and the agent is reset to another random non-goal position on the grid.


All agents are trained on pre-collected data samples of random-walk interactions and random resets in the environment. No reward signal is used to assist the pretraining stage. 
The encoder network architecture of $\phi$ is taken from the original paper, which is a 2-layer fully connected neural network with 32 hidden dimensions using the Tanh activation function. For quantization methods, the outputs of the encoder are quantized into a codeword in a codebook.
The training Loss of Markov abstraction is the weighted sum of the following three loss function
    \[\mathcal L_\text{Markov} = \alpha \mathcal L_\text{Inv} + \beta \mathcal L_{Ratio} + \eta \mathcal L_\text{Smooth}\]
We follow the original configuration to set the coefficients of the Inverse and Ratio loss to 1 and the smooth coefficient to 0 in this task.
In our experiment, the Continuous baseline is the original version of Markov abstraction, using similar setups as in \citep{allen2021learning}. 
For our quantization method, there is an additional Wasserstein loss from the objective in (\ref{eq:reconstruct_form_continuous_apd}),
we incorporate this quantization loss into the Markov loss as
\[\mathcal L = \mathcal L_\text{Markov} + \lambda \mathcal L_\text{quantize}\]
we set $\lambda = 100$ for all experiments with reinforcement learning through gridsearch the value of $\lambda\in \{0.1, 1, 10, 100\}$.
All experiments are repeated with 5 different seeds.


\end{document}